\documentclass[lettersize,journal]{IEEEtran}

\usepackage{mflogo,texnames}
\usepackage{enumerate}
\usepackage{float}
\usepackage{hyperref}       
\usepackage{url}            
\usepackage{booktabs}       
\usepackage{amsfonts}       
\usepackage{nicefrac}       
\usepackage{microtype}      
\usepackage{times}
\usepackage{xcolor}
\usepackage{soul}
\usepackage{CJKutf8}
\usepackage{lineno,hyperref}
\usepackage{enumerate}
\usepackage{amsfonts}       
\usepackage{graphicx}  
\usepackage{subfigure}
\usepackage{multirow}
\usepackage{amsmath}
\usepackage{float}
\usepackage{color}
\definecolor{c2}{HTML}{c89b40}
\usepackage[ruled]{algorithm2e}
\usepackage{framed}
\usepackage{url}
\usepackage{tabularray}
\UseTblrLibrary{diagbox}

\begin{document}
\begin{CJK}{UTF8}{gkai}
\title{LiDAttack: Robust Black-box Attack on LiDAR-based Object Detection}

\author{Jinyin~Chen,
        Danxin~Liao,
        Sheng~Xiang,
        Haibin~Zheng
        
\thanks{Manuscript received xx xx, xxxx; revised xx xx, xxxx. This research was supported by the Zhejiang Provincial Natural Science Foundation (No. LDQ23F020001), National Natural Science Foundation of China (No. 62072406). }
\IEEEcompsocitemizethanks{\IEEEcompsocthanksitem Jinyin Chen is with the Institute of Cyberspace Security, the College of Information Engineering, Zhejiang University of Technology, Hangzhou, 310023, China. (e-mail: chenjinyin@zjut.edu.cn). 
\IEEEcompsocthanksitem Danxin Liao is with the College of Information Engineering, Zhejiang University of Technology, Hangzhou 310023, China. (e-mail: 221122030373@zjut.edu.cn.).
\IEEEcompsocthanksitem Sheng Xiang is with the College of Information Engineering, Zhejiang University of Technology, Hangzhou, 310023, China. (e-mail:xiangsheng93@gmail.com).
\IEEEcompsocthanksitem Haibin Zheng is with the Institute of Cyberspace Security, Zhejiang University of Technology, Hangzhou, 310023, China. (e-mail: haibinzheng320@gmail.com).

\IEEEcompsocthanksitem Corresponding author: Sheng~Xiang.
}
}

\markboth{Journal of \LaTeX\ Class Files,~Vol.~14, No.~8, August~2021}%
{Shell \MakeLowercase{\textit{et al.}}: A Sample Article Using IEEEtran.cls for IEEE Journals}

\maketitle	
\begin{abstract}
Due to the significantly increasing number of autonomous driving systems incorporated with LiDAR sensors, the point cloud captured by LiDAR for object detection plays an indispensable role.  In particular, deep neural network (DNN) based point cloud object detection has achieved dominant performance compared with traditional methods. However, since DNN is vulnerable to carefully crafted adversarial examples, adversarial attack on LiDAR sensors have been extensively studied. In all, these attacks are challenged from three aspects: 
(i)~\emph{effective} - they're generally effective in white-box scenarios, but their performance will significantly degrade in black-box scenarios;
(ii)~\emph{robustness} - they are much less effective when exposed to the diverse changes in the real world (i.e., angle and distance changes);(iii)~\emph{concealing} - attacks will be defended since most of them overlook the stealthy goal of escaping detection.
To address these challenges, we introduce a robust black-box attack dubbed LiDAttack. It utilizes a genetic algorithm with a simulated annealing strategy to strictly limit the location and number of perturbation points, achieving a stealthy and effective attack. And it simulates scanning deviations, allowing it to adapt to dynamic changes in real world scenario variations.
Extensive experiments are conducted on 3 datasets (i.e., KITTI, nuScenes, and self-constructed data) with 3 dominant object detection models (i.e., PointRCNN, PointPillar, and PV-RCNN++). 
The results reveal the efficiency of the LiDAttack when targeting a wide range of object detection models, with an attack success rate (ASR) up to 90\%. To evaluate the robustness of LiDAttack, experiments at different distances and angles are conducted, and the results show that LiDAttack can maintain consistent performance over a certain range. The practicality of LiDAttack is further validated in the physical world both indoors and outdoors. The results show that LiDAttack-generated adversarial objects can still maintain effective attack in the physical world. In order to achieve a concealed attack, we limit the volume of the generated adversarial object to less than 0.1\% of the volume of the target object to achieve ASR up to 90\%. 
The code is available at \url{https://github.com/Cinderyl/LiDAttack.git}.

\end{abstract}
\begin{IEEEkeywords}
Deep learning, LiDAR-based, Object detection, Black-box adversarial attack, Physical attack, Defense.
\end{IEEEkeywords}

\IEEEpeerreviewmaketitle

\section{Introduction}
LiDAR sensor is an advanced technology that uses lasers to analyze the surrounding environment for wide-ranging areas, such as autonomous driving system. It calculates distance and 3D effects by measuring the time gap to fire a laser pulse, and returns it from the target. Among other things, the LiDAR sensor is capable of generating a point cloud for object detection\cite{lang2019pointpillars,yang2018pixor,shi2019pointrcnn}, enabling real-time monitoring of obstacles, pedestrians, and other vehicles around the target one. Thus, the autonomous driving system can make accurate decisions to ensure safe driving. 

From the perspective of representation learning strategies of the irregular laser point cloud, object detection can be broadly cast into three categories, i.e., point-based methods\cite{shi2019pointrcnn,yang2018ipod,yang2019std,pan20213d}, voxel-based methods\cite{lang2019pointpillars,zhou2018voxelnet,wang2020pillar,fan2022embracing}, and point-voxel based methods\cite{liu2019point,tang2020searching,he2020structure,miao2021pvgnet}.
Most of them rely on deep neural network (DNN) to process point cloud data due to the help of DNN's powerful extraction and learning capabilities. 
Not surprisingly, DNN-based point cloud object detection has achieved dominant performance.
However, recent research\cite{kurakin2018adversarial,szegedy2013intriguing} has revealed that DNN is susceptible to adversarial attack, in which imperceptibly perturbed examples will easily manipulate the DNN make incorrect predictions. 
Due to DNN's wide application in some safety-critical areas such as
 autonomous driving systems' LiDAR-based object detection\cite{zhou2023fastpillars,erabati2023li3detr}, the carefully crafted adversarial example of LiDAR point cloud will be a serious security threat. 

Furthermore, the vulnerability of LiDAR sensors towards physical attack will threat the autonomous driving system in a more direct manner. These attacks can be categorized into three groups, i.e., laser-based attack, object-based attack, and location-based attack. 
The laser-based attacks \cite{cao2019adversarial,cao2023you,hau2021object,sun2020towards,jin2023pla} are conducted by manipulating the signal detection system of a LiDAR sensor. Specifically, an attacker utilizes a photodiode to receive laser pulses from the LiDAR, and triggers the attack by activating a delay component, thereby simulating a real echo pulse. Generally, these attacks require prior knowledge of the victim's LiDAR laser emission frequency, so as to take the synchronization of the jammer and the attacked LiDAR into account. They fall under the category of white-box attack. Laser-based attack exhibits heightened sensitivity to the movement of the target object, resulting in a lack of robustness. For object-based attacks\cite{hau2021object,abdelfattah2021adversarial,abdelfattah2021towards,tu2020physically,cao2021invisible,tu2021exploring}, objects with adversarial shapes are proposed to attack LiDAR sensors. These adversarial objects are large and unusually shaped, resulting in inadequate \emph{concealment}.  
The location-based attack\cite{zhu2021can} aims to find the key confrontation location of the target object and fly the drone at the target location, but they are mostly effective in some scenarios, and the target is obvious and expensive. 
In all, the existing works are still challenged from three aspects. First, they lack \emph{concealability}, leading to easy detection. Second, they are susceptible to changes in the environment and targets, thus they lack adversarial \emph{robustness}. Third, most of these works 
are generally designed for specific scenario, while difficult to transfer to other applications, thus limited in \emph{generalizability}. 

To address these issues, we ask: ``~\emph{Can we implement a black-box attack to generate perturbation points to achieve a stealthy and robust physical attack with a high attack success rate (ASR) ?}" 

\begin{figure}
\centering
\includegraphics[width=0.5\textwidth]{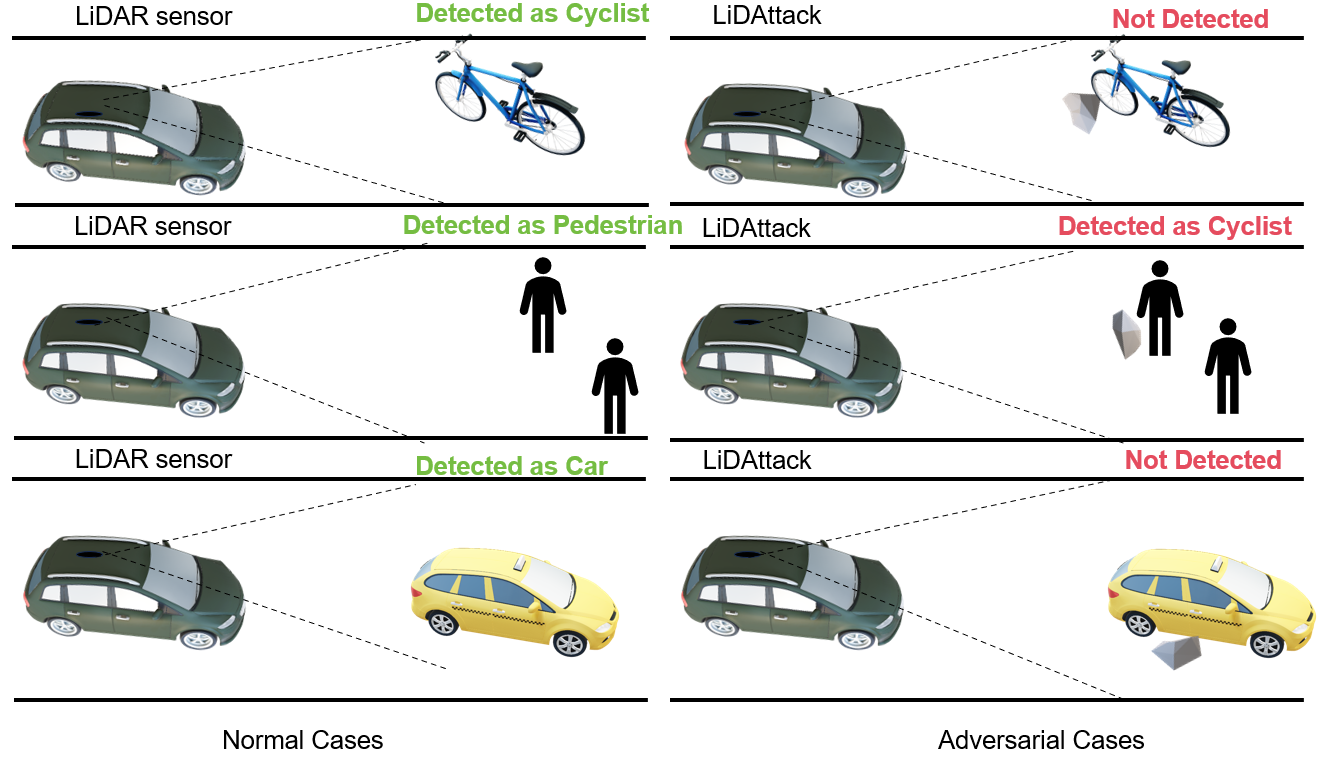}
\caption{Normal detection and adversarial scenarios. The left column indicates that the target object can be detected normally, and the right column indicates that the target object cannot be detected or misdetected by adding adversarial objects.}
\label{Fig1}
\end{figure}

To answer this question, a novel black-box adversarial attack is proposed based on genetic simulated annealing algorithm (GSA), denoted as \emph{LiDAttack}. Specifically, two scenarios are considered, i.e., hiding the target vehicle and detecting the target object as some others. In both scenarios, they can be successfully launched by placing a 3D printing optimally obtained from perturbation points near the target object. As shown in Fig. \ref{Fig1}, the left column indicates that the object detection model can correctly detect the target object in normal cases, and when the adversarial object is generated using LiDAttack and placed at the target location shown as the right column, object detection models fail to detect the target object or identify the target object as others, so that LiDAttack conducts the attack successfully.

In specific, the research questions are addressed as follows.

(i)~\emph{Can concealed adversarial objects be placed next to the target object for the purpose of a stealthy adversarial attack?} 
To address this challenge, the position of the perturbation point was carefully adjusted during the generation process to ensure that it remained within 0.2 meters away from the target object. This strategic placement is aimed at achieving effective adversarial attacks and sustained concealment.

(ii)~\emph{Whether the black-box attack used can achieve a high ASR?}
Trained adversarial examples are evaluated on three state-of-the-art point cloud object detection models, i.e., PointRCNN\cite{shi2019pointrcnn}, PointPillar\cite{lang2019pointpillars}, and PV-RCNN++\cite{shi2023pv}. The experiment proves that the ASR can reach more than 90\% on the object detection models.

(iii)~\emph{How robust is LiDAttack?}
The angular and distance deviations that exist in LiDAR are simulated in the digital domain, and the experimental results show that a certain range of deviations has only a small effect on attack performance. Furthermore, the robustness of LiDAttack as an effective physical attack in real-world scenarios is further verified.

In summary, the main contributions are as follows.
\begin{itemize}
  \item A novel black-box attack for point cloud object detection using GSA is proposed. LiDAttack achieves a highly stealthy attack by strictly controlling the distance between the perturbation point and the object as well as the distance between the perturbation points, so that they naturally blend into the background.

  \item An adaptive genetic algorithm is used to improve attack effectiveness and convergence speed. In the local search, simulated annealing algorithm is used to find a better solution, which involves accepting the probability of a worse solution to avoid falling into the local optimal solution trap. This approach combines the advantages of global and local search to find better solutions in a wider search space and improve the robustness of the algorithm.

  \item LiDAttack shows good performance on three datasets (i.e., KITTI, nuScenes, and self-constructed data) against wide-ranging detection models, covering point-based (e.g., PointRCNN), voxel-based (e.g., PointPillars), and hybrid point-voxel based (e.g., PV-RCNN++). LiDAttack can effectively break through the simple random sampling (SRS) defense mechanism. In addition, by applying LiDAttack combined with adversarial training, the robustness of the target detection models can be significantly enhanced. 

\end{itemize}

\section{Related Work}
We first briefly review deep learning for LiDAR-based object detection. Next, point cloud object detection and its potential attack and defense are fully explored.
\subsection{3D Object Detection based on LiDAR Point Cloud }
LiDAR-based object detection can be mainly categorized into point-based methods, voxel-based methods, and point-voxel based methods\cite{mao20233d}.

\textbf{Point-based 3D object detection.} Point-based 3D object detection\cite{shi2019pointrcnn,yang2018ipod,yang2019std,pan20213d} consists of two main basic components, i.e., sampling of the point cloud and feature learning. The point-based target detector first requires preprocessing of the point cloud, and then performs foreground or background point segmentation along with the learned high-dimensional feature points. This segmentation information is then used to bootstrap the anchor-free network to output a target candidate for each point. 

\textbf{Voxel-based 3D object detection.} Voxel-based point cloud object detection\cite{lang2019pointpillars,zhou2018voxelnet,wang2020pillar,fan2022embracing} converts irregular point cloud data into a regular form of voxels, each of which can be considered as a 3D pixel. A voxel is labeled as non-empty if a point cloud falls into that grid cell. Due to the sparsity of the point cloud data, most voxels are empty. Therefore, only non-empty voxels are stored and used for subsequent feature extraction and object detection. This approach can effectively handle large-scale point cloud data.

\textbf{Point-voxel based 3D object detection.} Point-voxel based methods use a hybrid architecture that utilizes points and voxels for 3D object detection\cite{liu2019point,tang2020searching,he2020structure,miao2021pvgnet,chen2019fast,shi2020pv,shi2023pv,mao2021pyramid,li2021lidar}.  These methods can be categorized into single-stage and two-stage detection frameworks. Single-stage point-voxel based 3D object detectors\cite{liu2019point,tang2020searching,he2020structure,miao2021pvgnet} organically combine point and voxel features with point-to-voxel and voxel-to-point transformations in the backbone network. In contrast, the two-stage point-voxel 3D object detector \cite{chen2019fast,shi2020pv,shi2023pv,mao2021pyramid,li2021lidar} employs a voxel-based detection framework in the first stage to generate a set of 3D object candidates, and in the second stage, it samples key points from the input point cloud and further optimizes the 3D candidates with point operators. 
\subsection{Point Cloud Adversarial Attack}
 Several recent studies have shown that networks that use point cloud data as input may be threatened by adversarial attack\cite{tu2020physically}. These attacks that use point clouds as inputs can be categorized into the following three main types, i.e., point shift-based attack\cite{xiang2019generating,liu2019extending,yang2019adversarial,lee2020shapeadv}, point add-based attack\cite{liu2020adversarial,kim2021minimal,arya2023adversarial}, and point drop-based attack\cite{zheng2019pointcloud,wicker2019robustness}. 

\textbf{Point shift-based attack.} It misleads the LiDAR-based object detection through small perturbations\cite{xiang2019generating,liu2019extending,yang2019adversarial,lee2020shapeadv}, which can be implemented by changing attributes such as the coordinates or reflection intensity of a point. 

\textbf{Point add-based attack.} It is another common digital domain attack designed to deceive the object detection system by adding extra points to the original point cloud\cite{liu2020adversarial,kim2021minimal,arya2023adversarial}. These extra points can be false obstacles, non-existent objects, or objects that do not match the real scene. 

\textbf{Point drop-based attack.} It works by selectively removing a number of points from the original point cloud in order to prevent the system from correctly sensing a specific object or region\cite{zheng2019pointcloud,wicker2019robustness}. 
Zheng et al\cite{zheng2019pointcloud} proposed an efficient method to construct a saliency map of a 3D point cloud to measure the contribution (importance) of each point to the model's prediction loss. Using the saliency map, they further standardized the point removal process. To explore the robustness of object detection models in adversarial environments, Matthew Wicker and Marta Kwiatkowska\cite{wicker2019robustness} demonstrated that critical point sets induced by latent spatial translations in 3D deep learning, for both point cloud and volume representations, expose weaknesses in adversarial occlusion attacks.
\subsection{Physical Adversarial Attack}
Physical attack on LiDAR-based object detection can be categorized into three types, i.e., laser-based attack\cite{cao2019adversarial,cao2023you,hau2021object,sun2020towards,jin2023pla} and object-based attack\cite{hau2021object,abdelfattah2021adversarial,abdelfattah2021towards,tu2020physically,cao2021invisible,tu2021exploring} and location-based attack\cite{zhu2021can}.

\textbf{Laser-based attack}. Through the attack, hacker can interfere with the victim's decision-making by firing lasers at them. Cao et al.\cite{cao2019adversarial} first found that a LiDAR-based perception deep learning model used in autonomous driving systems can cause the perception system to mistakenly detect a fake vehicle (one that does not actually exist) by injecting a small number of spoofed LiDAR points. Recently, Jin et al.\cite{jin2023pla} devised a method to convert point cloud coordinates into control signals and successfully implemented the attack on the physical domain.


\begin{table*}[ht]
\caption{The definition of symbols.}
\label{tab:symbols data}
\centering
\scalebox{0.85}{
\begin{tabular}{cc|r}
\hline \hline
\multicolumn{2}{c|}{\textbf{Symbol}}                                              & \textbf{Definition}                                                                          \\ \hline
\multicolumn{1}{c|}{\multirow{3}{*}{\textbf{Object Detection Model}}}         
& $D$, $D_{adv}$        & $D$ denotes input data, $D_{adv}$ denotes adversarial input data         \\\multicolumn{1}{c|}{} 
& $M$         & Object detection model   \\\multicolumn{1}{c|}{}
& $y$,  $y^*$ & $y$ denotes output labels corresponding to input data $D$,  $y^*$ denotes  ground truth labels  \\
\hline
\multicolumn{1}{c|}{\multirow{4}{*}{\textbf{Initialization}}}          
& $D_{init}$  & Population initialization input      \\\multicolumn{1}{c|}{}      &$\delta$     &Perturbation follows the Gaussian distribution \\\multicolumn{1}{c|}{} 
& $n$        & Population size        \\\multicolumn{1}{c|}{}    
& $N\left( \mu ,\sigma ^2 \right)$      & Gaussian distribution with mean $\mu$ and standard deviation $\sigma$ \\ 
\hline
\multicolumn{1}{c|}{\multirow{9}{*}{\textbf{Fitness Function}}}          
& $D_{i}^{t}$     & The $i$-th adversarial example of the $t$-th generation       \\\multicolumn{1}{c|}{}                                      
& $f\left( D_{i}^{t} \right)$         & The fitness function of $D_{i}^{t}$        \\\multicolumn{1}{c|}{}                                                           & $S\left( D_{i}^{t} \right)$         & Prediction score of $D_{i}^{t}$            \\\multicolumn{1}{c|}{}                                      
& $d_1$         & The distance between the perturbation point and the original target object
 \\\multicolumn{1}{c|}{}                                       
& $d_2$ & The distance between the perturbation points \\\multicolumn{1}{c|}{} 
& $\alpha _1, \alpha _2$ & The weight of the distance between the perturbation point and the original target object in the fitness function \\\multicolumn{1}{c|}{} 
& $\beta _1, \beta _2$ & The weight of the distance between the perturbation points in the fitness function \\
 \hline
\multicolumn{1}{c|}{\multirow{7}{*}{\textbf{Crossover Operator}}}          
& $p\left( D_{i}^{t} \right)$  & Probability that $D_{i}^{t}$ be selected           \\\multicolumn{1}{c|}{}                                       
& $q\left( D_{i}^{t} \right)$         & The cumulative probability of $D_{i}^{t}$  \\\multicolumn{1}{c|}{}                                       
& $k_c$     & Maximum crossover rate
\\\multicolumn{1}{c|}{}                                       
& $P_c$         & Crossover probability    \\\multicolumn{1}{c|}{}                & $f_{max}$         & Maximum fitness value of the population 
 \\\multicolumn{1}{c|}{}                                     
& $f'$         & The fitness value of an individual evaluated by the fitness function that needs to be crossover operation \\\multicolumn{1}{c|}{}              & $f_{avg}$      & The average fitness value \\
\hline
\multicolumn{1}{c|}{\multirow{4}{*}{\textbf{Mutation Operator}}}         
& $f$    & The fitness value of an individual evaluated by the fitness function that needs to be mutation operation \\\multicolumn{1}{c|}{}                       & $k_m$  & The maximum value of mutation rate    \\\multicolumn{1}{c|}{}         & $P_m$      & The probability of mutation \\ \multicolumn{1}{c|}{}               & $D'$      & Point cloud optimized by genetic algorithm \\
\hline
\multicolumn{1}{c|}{\multirow{4}{*}{\textbf{Simulated Annealing}}}         
& $Temp_0$        & The initial temperature \\\multicolumn{1}{c|}{}               & $Num$   & The number of annealing times  \\\multicolumn{1}{c|}{}                & $rand$      & Simulate the randomness of accepting new solutions \\ \multicolumn{1}{c|}{}                    & $\lambda $      & Cooling coefficient, which is used to control the rate at which the temperature decreases \\\hline
\hline
\end{tabular}}
\end{table*}

\begin{figure*}
\centering
\includegraphics[width=1\textwidth]{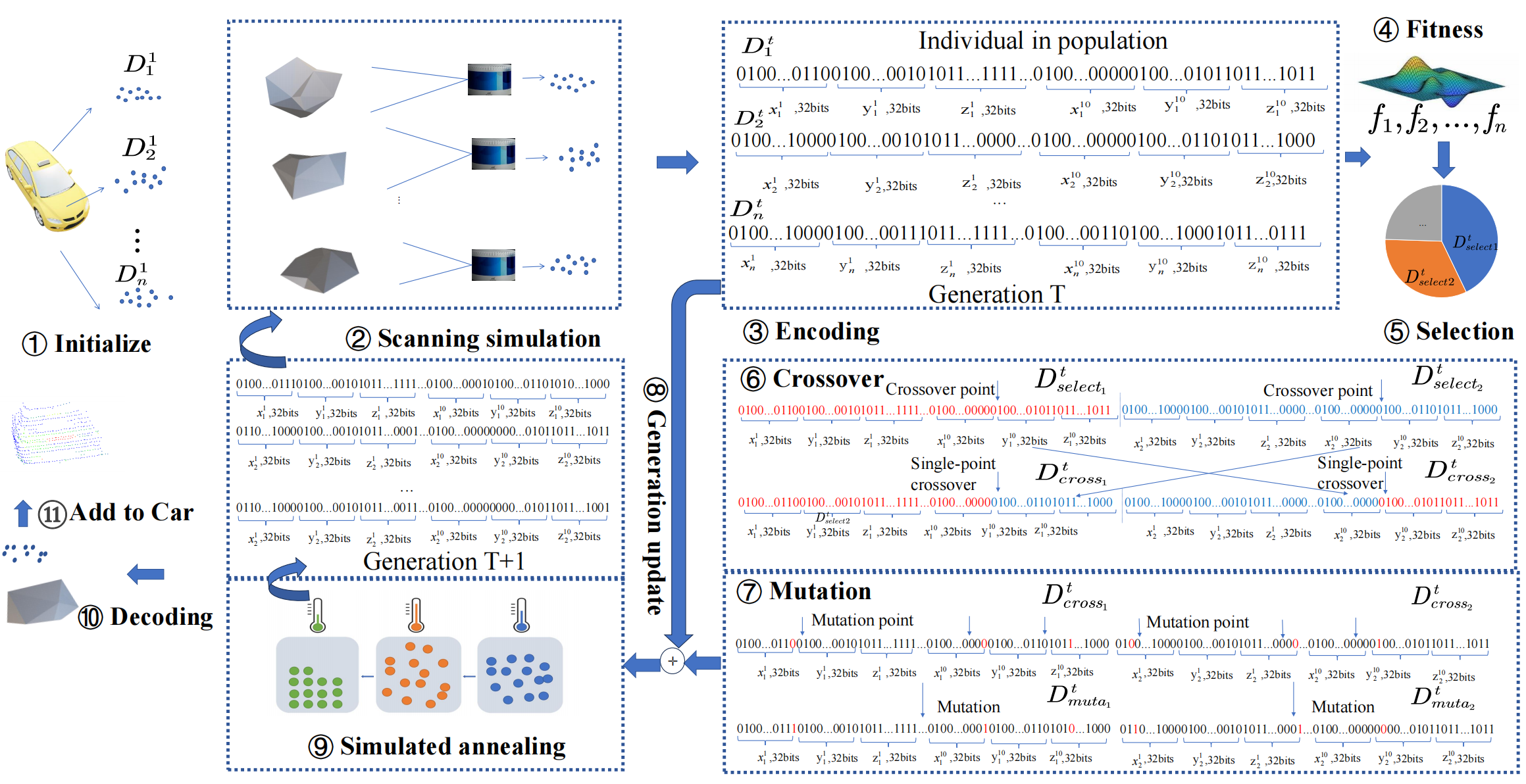}
\caption{The overall framework of LiDAttack.}
\label{Fig2}
\end{figure*}

\textbf{Object-based attack}. The attacker tries to breakdown a LiDAR sensor by using an object with an adversarial shape. Cao et al.\cite{cao2019objects} proposed an adversarial mesh based on a LiDAR renderer to generate an adversarial point cloud in order to generate a physically realizable attack and 3D printed the object to test it on a roadway. Tu et al.\cite{tu2020physically} learned to generate a generic and physically realizable adversarial object, demonstrating an attack to place an adversarial object on the roof of any target vehicle.

\textbf{Location-based attack}. Zhu et al.\cite{zhu2021can} proposed an innovative attack framework on which an attacker can identify key adversarial locations in physical space. By placing arbitrary objects with reflective surfaces around these locations, the attacker is able to easily spoof the LiDAR sensors.

\subsection{Defense for Point Cloud based Object Detection}
Adversarial defenses against point clouds generally fall into four categories, i.e., input transformations\cite{liu2019extending,SRS,yu2018pu,xiang2019generating}, data optimization, sensor-level optimization, and other defenses.

\textbf{Input transformation.} It is a preprocessing step before the deep model processes the point cloud, aiming to reduce the model's susceptibility to malicious attack and increase the difficulty of the attack by introducing transformations\cite{liu2019extending,SRS,yu2018pu,xiang2019generating}. Considering the relative prominence of the perturbed points of the input, 
Zhou et al.\cite{SRS} detected outliers and removed outliers by counting the Euclidean distance between each point and its k nearest neighbours. Xiang et al.\cite{xiang2019generating}, proposed a simple random sampling (SRS) to randomly remove a certain number of points from the input point cloud in order to reduce the influence of the perturbed points.

\textbf{Data optimization.} Data optimization refers to the optimization of training data to improve the robustness of deep models against adversarial attack\cite{liu2019extending,sun2021adversarially}. For example, data augmentation, adversarial training, etc. Liu et al.\cite{liu2019extending} first described adversarial training of point clouds. sun et al.\cite{sun2021adversarially} applied self-supervised learning to adversarial training of 3D point clouds. 

\textbf{Sensor-level optimization.} It is designed to improve the detection accuracy and robustness of an autonomous driving system by designing a structure for multi-sensor fusion. Many research efforts have been devoted to improving the performance of sensor fusion modules to enhance the safety and reliability of autonomous driving systems\cite{liang2018deep,chen2017multi,liang2019multi,du2017car}. 

\textbf{Other defenses.} Occlusion-aware structural anomaly detection (CARLO), and sequence view fusion (SVF) all designate the attack as a black-box emergence attack as proposed in Sun's work\cite{sun2020towards} and defend against the attack by removing dubious points from the input point cloud or deleting suspicious objects from the final prediction. Xiao et al.\cite{xiao2023exorcising} proposed a novel defense module called local objectness predictor (LOP), which detects and removes false obstacles in LiDAR target detectors by learning to predict the truthfulness of local parts of an obstacle, thus effectively defending against emergent attack in autonomous driving systems.

\section{Preliminaries}
This section introduces the attack scenario for object detection, the object detection model, and the definitions of adversarial attack. For convenience, the definitions of some important symbols are listed in Table \ref{tab:symbols data}.

\subsection{Object Detection Model}
The principle of LiDAR-based point cloud object detection is to process and analyze the point cloud data collected by LiDAR to achieve the detection and identification of target objects. In this case, the point cloud data is composed of distance values that are measured by the LiDAR to measure the flight time of the emitted laser beam and calculated. The specific detection principle is as follows.

Given a deep learning model $M$, input data $D$ and corresponding ground truth labels $y^*$. The goal of the object detection model is to output labels that are the same as the ground truth labels after inputting data $D$, i.e., $y=M\left( D \right)$, $y =y^*$.

\subsection{Adversarial Attack}
The goal of an attacker against an autonomous driving system is usually to interfere with or disrupt the proper functioning of the object detection model in the system, rendering the object detection ineffective, which can lead to unsafe driving or navigational behaviors. The specific attack principle is as follows.

Given a deep learning model $M$, input data $D$, and corresponding ground truth labels $y^*$, the goal of the adversarial attack is to find an adversarial input $D_{adv}$ such that the model output is inconsistent with the ground truth labeling, i.e., $y=M\left( D_{adv} \right)$, $y \ne y^*$. The adversarial input $D_{adv}$ is usually obtained by solving an optimization problem. In physically realizable attack, the adversarial input $D_{adv}$ is usually implemented by making modifications in the original input $D$. 
\section{Methodology}

\subsection{Framework}
A black-box attack named LiDAttack is proposed, which aims to achieve an attack on the target detection model by optimizing the location of perturbation points to manipulate it to make incorrect detection. Specifically, LiDAttack is designed based on the GSA, which combines the global search capability of the genetic algorithm and the local search capability of the simulated annealing algorithm, thus it is able to efficiently generate covert and effective perturbation points. 

Fig. \ref{Fig2} illustrates the main framework of LiDAttack. In the initial stage, the system randomly generates perturbation points and adds them to the target point cloud. Then, these perturbation points are modeled and scanned by a simulated LiDAR to obtain the corresponding simulation results. Afterwards, the coordinates of the perturbed points are encoded as chromosomes and a fitness function is defined to evaluate the effectiveness of these perturbed points. Subsequently, a new generation of perturbations is generated using the operations of selection, crossover, mutation, generation update, and simulated annealing in the simulated annealing algorithm. This process continues until the termination condition is met, at which point the system outputs the adversarial example with the approximate optimal fitness function value. 

\subsection{Technical Designs}
In LiDAttack, global search is mainly realized by genetic operations, i.e., searching for optimal solutions by selection, crossover, and mutation operations. Adaptive genetic algorithm\cite{srinivas1994adaptive} is used to improve convergence accuracy and speed. The algorithm automatically adjusts the crossover and mutation probabilities according to the changes in the fitness function values. Initially, a larger probability is selected to maintain diversity, and later it is gradually adjusted down to avoid destroying the optimal solution. The genetic algorithm has a strong global search capability and is able to perform extensive searches in the entire solution space, thus avoiding premature convergence to a local optimum. While the simulated annealing algorithm is good at performing local search, it helps to jump out of the local optimal solution by accepting the worse solution with a certain probability, which enhances the algorithm's local search ability. The specific process is shown as follows.
\subsubsection{\textbf{Initialize}}

The initial solution is obtained through random initialization, i.e., generating random Gaussian perturbation points within a specified range, Gaussian perturbation can generate random numbers with different means and standard deviations, which helps introduce diversity into the initial population. The corresponding initial adversarial example can be expressed as $D_{init}=D+\delta$, where $\delta \sim N\left( \mu ,\sigma ^2 \right)$. To improve the diversity of initial perturbations, different types of initial perturbations are generated for different numbers of perturbation points in this paper.

\subsubsection{\textbf{Scanning simulation}}
Considering that the uncertainty in the point cloud reconstruction process will weaken the ASR in the real environment, the initial adversarial point cloud is obtained by creating physical objects at the beginning of the attack and scanning them with simulated LiDAR. Before each time perturbed points need to be fed into the object detection model to obtain a confidence score, a reconstruction of the physical object has to be performed and then the point cloud is obtained by simulating a LiDAR scan. Simulated LiDAR scanning not only ensures that the best perturbation points are obtained, but also guarantees the accuracy of the physical object.

The simulation of the laser beam is divided into two directions, i.e., horizontal and vertical. 
During the simulation, the direction of the laser beam is generated for each horizontal and vertical angle, and its starting point is set to the position of the laser source. The points on the model surface are then obtained by calculating the intersections with the triangular faces. The specific details of the simulated LiDAR scanning process are shown in Algorithm~\ref{algorithm_1}.  The 3D object corresponding to the final obtained perturbation point is printed.
\begin{algorithm}[ht]
\label{algorithm_1}
\caption{Scanning simulation.}
\LinesNumbered
\KwIn{ Number of laser beams $num\_{beams}$, vertical angle of each laser beam $A$, horizontal angular resolution $\gamma $, object information including vertex $vectors\in R^3$, starting position of the laser beam $R(r_x,r_y,r_z)$}
\KwOut{ Get the scanned points $scan\_{points}$} 
Generate equally spaced horizontal angle numbers based on horizontal angle resolution $H=\left[ 0,\gamma,...,360 \right]$.\\
\For {$h$ in $H$}
{
        \For {$a$ in $A$} 
        {
       \textcolor{brown}{ $\triangleright$ the direction of the laser beam $D_i$}
$D_i=\left\{ \cos a\cdot \cos\text{h,}\cos a\cdot \sin\text{h,}\sin a \right\} 
$\\

         \For {$triangle$ in $vectors$} 
         {
          
         \textcolor{brown}{ $\triangleright$ $v_0,v_1,v_2$ represents each of the three vertices of the triangle}
         $v_0,v_1,v_2=triangle$\\
         \textcolor{brown}{ $\triangleright$ $n$ denotes the normal vector of the triangle}\\
         $n=\left( v_1-v_0 \right) \times \left( v_2-v_0 \right)$\\
          \uIf{n==0}
          {
          continue;
          }
          \Else
          {
           \textcolor{brown}{ $\triangleright$ intersection of the laser beam with the surface}\\
          $point=R+\frac{\left( v_0-R \right) .n}{D_i\cdot n}$
          }
           \textcolor{brown}{ $\triangleright$ check if the intersection is inside the triangle}\\
          $u=\frac{\left( v_1-v_0 \right) \cdot \left( point-v_0 \right)}{\left( v_1-v_0 \right) \cdot \left( v_1-v_0 \right)}$\\
          $v=\frac{\left( v_2-v_1 \right) \cdot \left( point-v_1 \right)}{\left( v_2-v_1 \right) \cdot \left( v_2-v_1 \right)}$\\
          $w=1-u-v$

          \If{$0\le u,v,w\le 1$}
          {
           \textcolor{brown}{ $\triangleright$ combine intersecting points into the point cloud}\\
          $scan\_{points}$$ \gets$ $point$
          }
          
         }
     
        }
}
\Return $scan\_{points}$
\end{algorithm}

\begin{figure}
\centering
\includegraphics[width=0.5\textwidth]{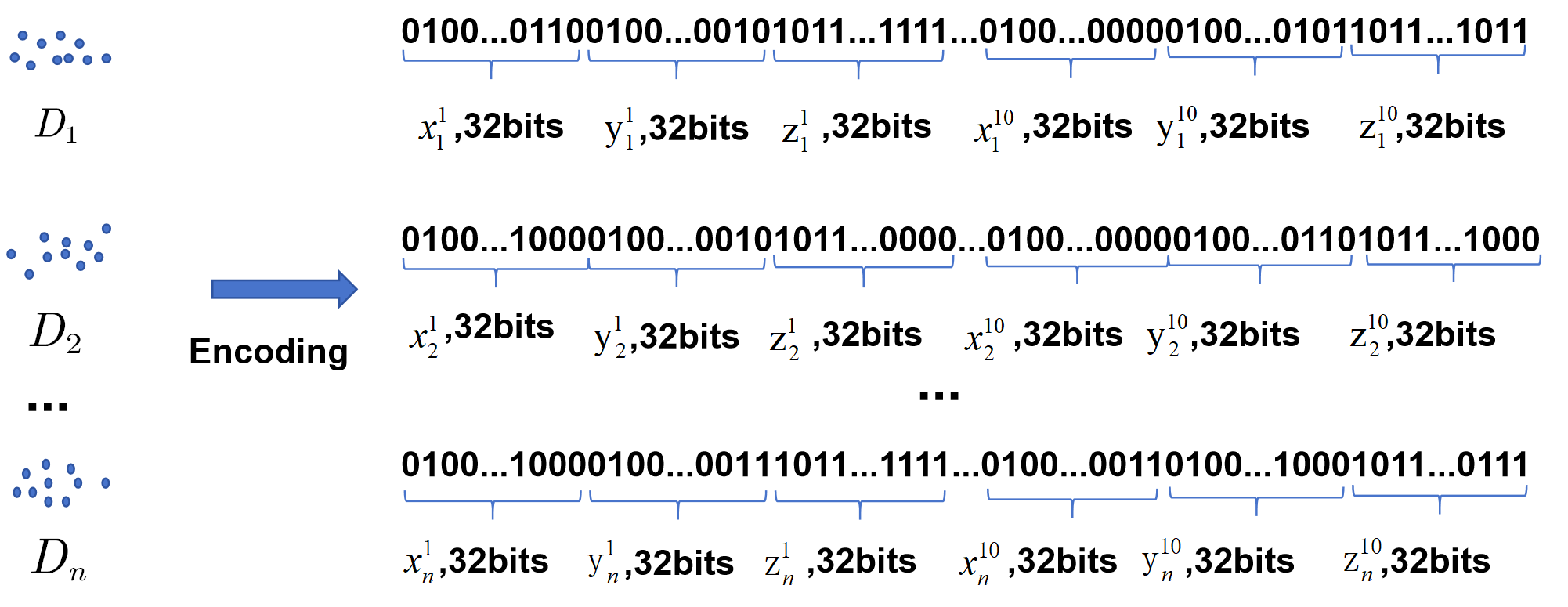}
\caption{Encoding process in LiDAttack.}
\label{Fig3}
\end{figure}

\begin{figure}
\centering
\includegraphics[width=0.48\textwidth]{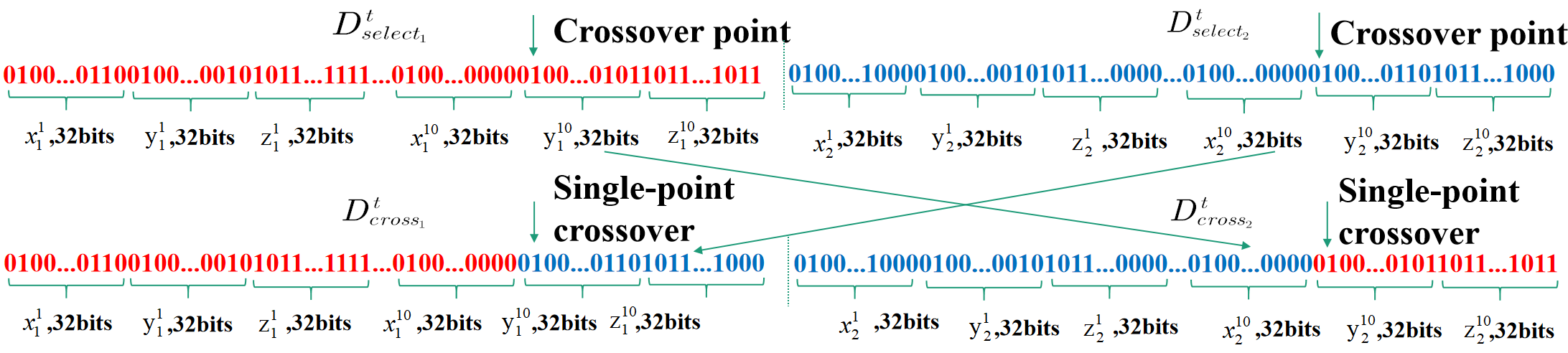}
\caption{Crossover process in LiDAttack.}
\label{crossover}
\end{figure}

\begin{figure}
\centering
\includegraphics[width=0.48\textwidth]{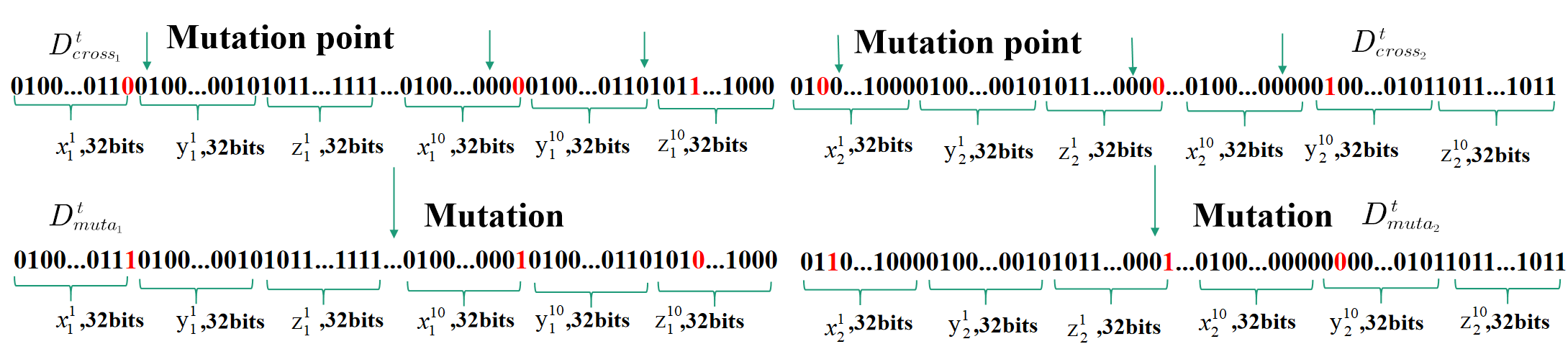}
\caption{Mutation process in LiDAttack.}
\label{Mutation}
\end{figure}

\subsubsection{\textbf{Encoding}}
 The method of encoding directly affects the operation of genetic operators such as crossover operators and mutation operators of genetic algorithms, and therefore largely determines the efficiency of genetic evolution. Genetic algorithms have a variety of coding methods\cite{deb1999introduction}, the most common of which is binary encoding. Binary coding is chosen to convert the coordinate value of each point into a binary string, and then multiple binary strings are connected to form a chromosome. The process of realization is as follows.

Taking the example of a population size, Fig. \ref{Fig3} illustrates the process of encoding. In this example, each individual consists of 10 perturbation points. Each floating point number in the coordinates of the perturbation point is encoded using a 32-bit binary representation. The reason for choosing this encoding method is its relatively compact nature, which occupies less storage space and helps to reduce the data storage requirements. In addition, this encoding method also facilitates the subsequent crossover and mutation operations of the genetic algorithm.


\subsubsection{\textbf{Fitness function}}
The fitness function is used in genetic algorithms to evaluate the superiority or inferiority of an individual or solution, representing its ability to compete for survival in its environment. In each iteration, the fitness function evaluates the individuals, and those with high scores are more likely to be selected for reproduction. Effective construction and utilization of the fitness function will determine the search direction and evolutionary outcome of the genetic algorithm, affecting the algorithm's performance and scope of application.

The adversarial example generated possesses the following two characteristics. First, it is relatively close to the original target point cloud, with only some subtle perturbations that are almost indistinguishable to the naked eye; second, the adversarial example can be incorrectly recognized by the target model as other classes or not recognized with a high confidence level. Therefore, the fitness function in the perturbation optimization process is related to the attack effect and the size of the perturbation. The fitness function is calculated as follows.

(i) The fitness function is defined as follows when the original object can still be recognized correctly after adding the perturbation point.
\begin{equation}
f_{1} \left( D_{i}^{t} \right) =1-S\left( D_{i}^{t} \right)
\label{fit1}
\end{equation}
where $f_{1} \left( D_{i}^{t} \right)$ denotes the fitness function of $D_{i}^{t}$ when the object detection model still recognises the object after adding the adversarial perturbation as the original object, $S\left( D_{i}^{t} \right)$ denotes the prediction score of $D_{i}^{t}$. The purpose of this formulation is to achieve a faster attack by ignoring the size of the perturbation until the attack is successful, which leads to errors in the object detection model after the inclusion of the perturbation point.

(ii) If the object detection model does not recognize the original object, the attack is successful. While maintaining a low confidence level it should make the perturbation points as close as possible to the surface of the object, which can be used to achieve a stealthy effect when physically assigning the shape. In addition, the distance between the perturbation points should be as close as possible for physical realization.
\begin{equation}
f_{2} \left( D_{i}^{t}\right)=\left( 1-S\left( D_{i}^{t}\right) \right) +\alpha_1 \left( \frac{1}{1+d_1} \right) +\beta_1 \left( \frac{1}{1+d_2} \right)
\label{fit2}
\end{equation}
where $f_{2} \left( D_{i}^{t} \right)$ denotes the fitness function of $D_{i}^{t}$ when the object detection model does not recognize the original object, $d_1$ denotes the distance between the perturbation point and the original target object, expressed as the Chamfer distance, and $d_2$ denotes the distance between the perturbation points, expressed as the Euclidean distance. $\alpha_1 $ denotes the weight of the distance between the perturbation point and the original target object in the fitness function, and $\beta_1$ denotes the weight of the distance between the perturbation points in the fitness function.

(iii) If the object detection model recognizes the original object as another class, the attack can be considered successful.In this case, a higher prediction score would be advantageous. In addition, the distance $d_1$ and $d_2$ need to be further optimized.
\begin{equation}
f_{3} \left( D_{i}^{t} \right) =S\left( D_{i}^{t} \right) +\alpha_2 \left( \frac{1}{1+d_1} \right) +\beta_2 \left( \frac{1}{1+d_2} \right)
\label{fit3}
\end{equation}
where $f_{3} \left( D_{i}^{t} \right)$ denotes the fitness function of $D_{i}^{t}$ when the object detection model recognizes the original object as another type of object, $\alpha_2 $ denotes the weight of the distance between the perturbation point and the original target object in the fitness function, and $\beta_2$ denotes the weight of the distance between the perturbation points in the fitness function.

\subsubsection{\textbf{Selection}}
The selection operation can determine the search direction, and improve global convergence and computational efficiency. In genetic algorithms, the principle of ``survival of the fittest and selection of the fittest" allows individuals with higher fitness to have a greater chance of passing on to the next generation. In this way, it helps to prevent the search process deviating from the direction of the optimal solution. 

Examples are selected using the roulette wheel selection. For a given example $D_{i}^{t}$, its probability of being selected, $p\left( D_{i}^{t} \right) $, can be obtained from Equation \ref{selecte}, and the interval $\left( q\left( D_{i-1}^{t} \right),q\left( D_{i}^{t} \right) \right) $ of being selected in the model can be calculated according to Equation \ref{range}. 
\begin{equation}
p\left( D_{i}^{t} \right) =\frac{f\left( D_{i}^{t} \right)}{\sum_{j=1}^n{f }\left( D_{j}^{t} \right)}
\label{selecte}
\end{equation}

\begin{equation}
q\left( D_{i}^{t} \right) =\sum_{j=1}^t{p}\left( D_{j}^{t} \right)
\label{range}
\end{equation}

\subsubsection{\textbf{Crossover}}
Genetic algorithms maintain population diversity through crossover operators. Single-point crossover is utilized. As shown in Fig. \ref{crossover}, it works by randomly selecting a crossover point and then swapping the partial genes of the two parent chromosomes after that point.

An adaptive genetic algorithm is employed, which can dynamically adjust the probabilities and parameters of operations such as crossover and mutation according to the fitness of the population and the characteristics of the problem, so as to search more flexibly in different stages and situations. The formula for the example crossover probability is as follows.
\begin{equation}
P_c=\left\{ \begin{array}{l}
	\frac{k_c\left( f_{\max}-f' \right)}{f_{\max}-f_{avg}},f'\ge f_{avg}\\
	k_c,f'<f_{avg}\\
\end{array} \right. 
\label{crosse1}
\end{equation}
where $f'$ represents the fitness value of an individual evaluated by the fitness function that needs to be
crossover operation, $f_{avg}$ represents the average fitness value of the population, and $f_{\max}$ represents the maximum fitness value of the population, and $k_c$ is the maximum crossover rate, $0<k_c <1$. 

\subsubsection{\textbf{Mutation}}
The main purpose of the mutation operation is to maintain the diversity of the population by introducing stochastic variations to prevent the algorithm from prematurely converging to a locally optimal solution. 

As shown in Fig. \ref{Mutation}, multipoint mutations are utilized, which are random changes made to multiple genes at the same time in an individual's chromosomal coding strings to generate new genotypes. Multipoint mutation is opposed to single-point mutation, which changes values at one gene at a time. In contrast, multipoint variation can affect multiple features at once, a mechanism that not only increases the diversity of the population, but also enhances the algorithm's ability to explore the unknown solution space. By introducing multiple change points, multipoint mutation gives the algorithm a higher chance of escaping the local optimality trap and thus potentially discovering new, superior solutions. The probability of variation is calculated as follows.
\begin{equation}
P_m=\left\{ \begin{array}{l}
	\frac{k_m\left( f_{\max}-f \right)}{f_{\max}-f_{avg}},f\ge f_{avg}\\
	k_m,f<f_{avg}\\
\end{array} \right. 
\label{mutation1}
\end{equation}
where $f$ represents the fitness value of an individual evaluated by the fitness function that needs to be mutation operation. $k_m$ is the maximum value of mutation rate, $0<k_m<k_c<1$.
\subsubsection{\textbf{Generation update}}
Generation updating is the process of updating the individuals in a population so that the algorithm can perform a targeted search in the solution space to find a better solution to the problem. The elite strategy is employed to update both the parent and child generations, i.e., the elite individuals from the parent generation are combined with the highest-performing individuals from the child generation to form a new population, $D^{'}$. $D^{'}$ will subsequently be utilized in simulated annealing.
\begin{algorithm}[]
\label{algorithm_2}
\caption{LiDAttack attack.}
\LinesNumbered
\KwIn{ Initial temperature $Temp_0$, number of annealing times $Num$,object detection model $M$, point cloud $D$, number of iterations $T$, population size $N$, $k_c$, $k_m$, number of perturbation points $n_0$, fitness function $F$.}
\KwOut{The perturbation point $D_{opt}$ with the greatest fitness.} 
Extract the point cloud of the target object $D_{target}$.\\
\For {$population=1,2,...,N$ }
{
\textcolor{brown}{ $\triangleright$ The population is initialized by adding small random Gaussian $\delta$ noise to the point cloud Dtarget of the target object, and randomly sampling $n_0$ perturbation points.}\\
 $\delta \sim N\left( \mu ,\sigma ^2 \right)$\\
3D reconstruction and LiDAR simulation scanning using Algorithm \ref{algorithm_2} to get the input perturbation points $\delta'$.
}
        \For {t=1,2,…,T} 
        {
            \For{t=1,2,...,N}{
             \textcolor{brown}{ $\triangleright$  Scoring of populations according to the fitness function $F$}\\
                \uIf{$y_1=y_*$}
                {
                 $f \left( D_{i}^{t} \right) =1-S\left( D_{i}^{t} \right)$
                 }
                  \uElseIf{$y_1\ne y_*$}
                  {
                 $ f \left( D_{i}^{t} \right) =S\left( D_{i}^{t} \right) +\alpha_2 \left( \frac{1}{1+d_1} \right) +\beta_2 \left( \frac{1}{1+d_2} \right)$
                  }
                  \Else{
                  $f \left( D_{i}^{t}\right)=\left( 1-S\left( D_{i}^{t}\right) \right) +\alpha_1 \left( \frac{1}{1+d_1} \right) +\beta_1 \left( \frac{1}{1+d_2} \right)$
                  }
                   \textcolor{brown}{ $\triangleright$  Selection}\\
                    $p\left( D_{i}^{t} \right) =\frac{f\left( D_{i}^{t} \right)}{\sum_{j=1}^n{f }\left( D_{j}^{t} \right)}$,
             $q\left( D_{i}^{t} \right) =\sum_{j=1}^t{p}\left( D_{j}^{t} \right)$\\
               \textcolor{brown}{ $\triangleright$  Crossover}\\
               \uIf{$rand\left( 0,1 \right) <P_c$}
               {
               $D_{cross_{1}}^{t}=D_{select_{1}}^{t}*A+D_{select_{2}}^{t}\left( 1-A \right) $
               $D_{cross_{2}}^{t}=D_{select_{1}}^{t}*\left( 1-A \right) +y_{select_{2}}^{t}A$
               }\Else{
               $D_{cross_{1}}^{t}=y_{selec_{1}}^{t}$,
               $D_{cross_{2}}^{t}=y_{selec_{2}}^{t}$\\
               }
             \textcolor{brown}{ $\triangleright$  Mutation}\\
               \uIf{$rand\left( 0,1 \right) <P_m$}{
               $y_{muta}^{t}=y_{cross}*C$
               }\Else{
                $y_{muta}^{t}=y_{cross}$
               }
               
            }
            Population update from offspring and parents using an elite update strategy.\\
          Localized search reference part IV.C\\
        }
        
\Return the individual $D_{adv}$ with the largest fitness. 
\end{algorithm}
\subsubsection{\textbf{Simulated annealing}}
The simulated annealing algorithm is used to update the output of each round of the genetic algorithm as follows.

(i) Given the initial temperature $Temp_0$, the number of annealing times $Num$ and the population $D'$ obtained from the genetic algorithm, the fitness value $f\left( D_{i}^{'} \right) $ is calculated for each individual.

(ii) Generate a random perturbation $\delta $, obtain a new individual $D_{i}^{''}=D_{i}^{'}+\delta $, calculate the fitness of the individual $f\left( D_{i}^{''} \right) $, and the difference between the fitness values $\varDelta f=f\left( D_{i}^{'} \right) -f\left( D_{i}^{''} \right) 
$.

(iii) If $\varDelta f\ge 0$, the new individual is accepted as the initial individual for the next simulation; otherwise, $\varDelta f<0$, then calculate the probability of receiving the new individual: $P\left( \varDelta f \right) =e^{\left( \varDelta f\cdot Temp \right)}
$, generate a uniform random number $rand$ on the interval [0,1], if $P\left( \varDelta f \right) \ge rand$, then accept the new individual as the initial individual for the next simulation, otherwise discard the new individual and still take the original individual as the initial individual for the next simulation.

(iv) $Temp\left( Num \right) =\lambda Temp\left( Num \right)$, $Num=Num+1$,where $\lambda$ is a positive number less than 1. If the stopping condition is satisfied, the computation stops, otherwise, return (ii).

\subsubsection{\textbf{Decoding}}
The decoding operation in a genetic algorithm is the process of converting a chromosome (i.e., a binary string) into an individual solution. In binary coding, each gene corresponds to a binary bit, so the decoding operation is the conversion of the binary string into the corresponding values in the order of the genes. In other words, decoding is the inverse process of encoding. The process of realization is as follows.

In the decoding process, the binary string representation of each floating point number is converted to an integer. Subsequently, by converting these integers to hexadecimal strings with appropriate padding and formatting operations, IEEE 754 compliant hexadecimal representations are obtained. Finally, by using a big-endian byte order, the hexadecimal strings are parsed into 32-bit floating point numbers. The purpose of this sequence of steps is to successfully reduce a floating-point number in binary form to a computable floating-point format. This conversion process is commonly used in data processing to ensure the correctness and consistency of data. LiDAttack details are shown in Algorithm~\ref{algorithm_2}.

\section{Experimental Setting}
\textbf{Platform.} Intel XEON6420 2.6GHzX128C (CPU), Tesla V100 32GiG (GPU), 16GB DDR4 RECC2666 (memory), Ubuntu 16.04 (OS), Python 3.7. In this experiment, 3D printing was utilized to construct the necessary adversarial objects. In the physical experiments, space data acquisition relied on the Mid-40 lidar sensor from the Livox.

\textbf{Victim models.} For the experiments in the digital world, PointRCNN\cite{shi2019pointrcnn}, PointPillar\cite{lang2019pointpillars}, and PV-RCNN++\cite{shi2023pv} are chosen as the target models. PointRCNN is an advanced point cloud object detection framework that improves the detection accuracy in complex scenes through a two-stage process of first extracting candidate regions, followed by fine classification and localization. PointPillar is a real-time 3D object detection that efficiently handles large-scale point clouds and is suitable for rapid dynamic environment analysis by converting point cloud data into columns perpendicular to the ground and applying a 2D convolutional network. PV-RCNN++ is an efficient detection network developed on the basis of PointRCNN, which employs multi-scale feature fusion and an improved regional proposal network to enhance the detection of targets of different sizes and improve the model robustness. The performance on different datasets  can be found in Table \ref{datamAP}.

\textbf{Dataset.} For point cloud data, the KITTI, nuScenes, and self-constructed data are used.
\begin{itemize}
\item KITTI\cite{geiger2012we}. KITTI is widely used to train many state-of-the-art object detection models. In our experiments, from the KITTI dataset, 3000 LiDAR examples were extracted and used as a clean training set, which is inaccessible to the attacker. The KITTI dataset categorizes the recognition difficulty of the target object into ``Easy", ``Moderate", and ``Hard". The recognition difficulty is defined according to whether the labeled box is occluded, the degree of occlusion and the height of the box. The focus is on three categories of objects: ``Car", ``Pedestrian" and ``Cyclist". 

\item nuScenes\cite{caesar2020nuscenes}. nuScenes is a large multi-model autonomous driving dataset with a full 360° field of view in a variety of challenging urban driving scenarios. It includes 1000 scenarios collected in Boston and Singapore under different weather conditions. Results are reported from the entire validation set.

\item Self-constructed data. The self-constructed dataset is derived from more than 50 outdoor scenes and 50 indoor scenes, captured using a camera and Livox LiDAR, with a total of 1000 instances. Results are reported from the entire validation set. Some of the self-constructed dataset is available at \url{https://github.com/Cinderyl/self-constructed-data}.
\end{itemize}

\textbf{Evaluation metrics.} Attack success rate (ASR) is utilized as a metric, which is the ratio of the number of successful attacks on an object detector to the total number of attacks that have been performed. In the field of object detection, models usually set a default confidence score threshold with the purpose of excluding detections that have low predictive confidence. In this study, the ASR is measured based on the default confidence score thresholds of the three target models. The specific evaluation criteria are as follows: when the label predicted by the model matches the real label and the prediction confidence of that label is lower than the default threshold, the attack is considered successful; if the label predicted by the model does not match the real label and the prediction confidence exceeds the default threshold, the attack is likewise recorded as successful. Mean Average Precision (mAP) is the average of the average precision (AP) on multiple categories. For each category, AP is a measure of the area under the Precision-Recall curve for the category. In addition, the misclassification rate is used as an evaluation metric, i.e., the ratio of the number of misclassified examples to the total number of examples.

\begin{table}
\centering
\caption{mAP values of three target detection models of PointRCNN,PointPillar,PV-RCNN++ on three datasets of KITTI,nuScenes,self-constructed data.}
\label{selft}
\begin{tblr}{
  cells = {c},
  vline{2} = {-}{},
  hline{1-2,5} = {-}{},
}
\diagbox{Model}{Dataset} & KITTI   & NuScenes & Self-constructed data \\
PointRCNN                  & 68.41\% & 39.23\%  & 73.26\%               \\
PointPillar                & 64.08\% & 46.58\%  & 76.71\%               \\
PV-RCNN++                  & 76.68\% & 67.02\%  & 80.40\%               
\end{tblr}
\label{datamAP}
\end{table}

\textbf{Algorithms settings.} The settings for each parameter of the LiDAttack attack are shown in Table \ref{Algorithms Settings}.

\begin{table}
\centering
\caption{Algorithm parameter setting.}
\begin{tabular}{c|c} 
\toprule
Parameters                                   & Value                 \\ 
\hline
Population size $n$                             & 20                    \\
Population initialization standard deviation $\sigma ^2$ & 0.01                  \\
The number of iterations of evolution  $T$      & 1000                  \\
Crossover, mutation probability  $P_c$, $P_m$            & Adaptive probability  \\
$k_c $                                          & 1                     \\
$k_m$                                        & 0.5                   \\
Initial temperature $Temp_0$                         & 300                   \\
The number of annealing times $Num$ & 500                   \\
Cooling rate $\lambda$                                & 0.98                  \\
Termination temperature                      & 1.4                   \\
\bottomrule
\end{tabular}
\label{Algorithms Settings}
\end{table}

\section{Experimental Results and Analysis}
In this section, the performance of LiDAttack is evaluated in terms of attack effectiveness, generalisability, and practicality. The following research questions (RQs) are posed.

\textbf{RQ1. Effectiveness}: How effective is LiDAttack against object detection model?

\textbf{RQ2. Robustness}: How about LiDAttack's robustness?

\textbf{RQ3. Practicality}: How does the physical practicality of LiDAttack?

\textbf{RQ4. Adversarial Training}: How to defend against LiDAttack and improve model robustness?

By answering these questions, a comprehensive understanding of the strengths and limitations of LiDAttack, as well as its potential threats in the real world, can be gained.
\subsection{\textbf{RQ1.} How effective is LiDAttack against object detection model?}

\begin{figure}
	\centering
	\subfigure[different target objects]{
		\includegraphics[width=0.46\linewidth]{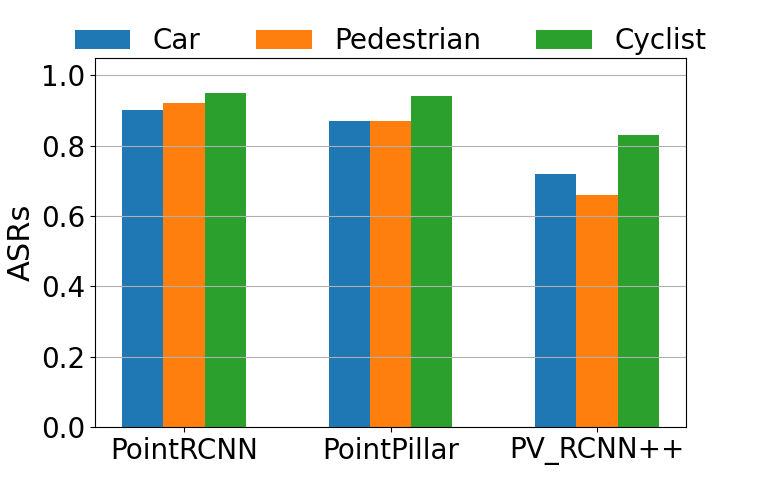}
	}
        \subfigure[different perturbation points]{
		\includegraphics[width=0.46\linewidth]{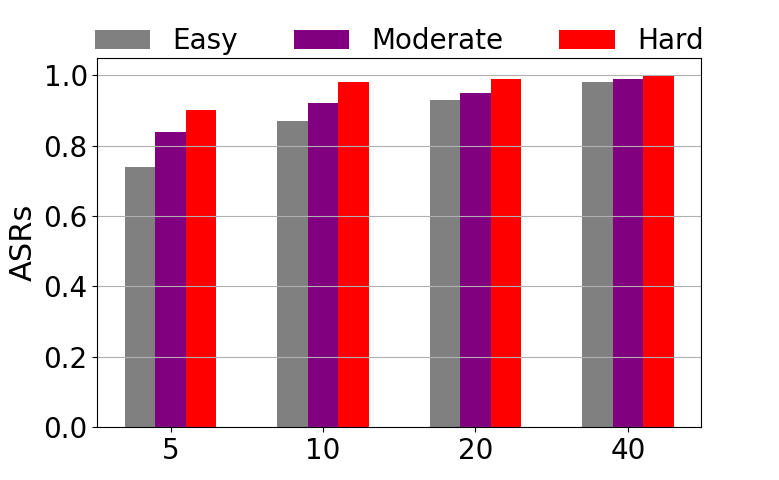}
	}
	
	\caption{ASR of LiDAttack attack under different target objects and perturbation points. (a) ASR of LiDAttack varies with different target objects when the perturbation point is set to 10. (b) ASR of LiDAttack attack under different perturbation points. }
\label{different object and points}
\end{figure}

When reporting the results, we focus on the following aspects, i.e.,  different types of objects, the number of perturbation points and different recognition difficulty. The evaluation results are shown in Table \ref{2} and Fig. \ref{different object and points}.

\begin{table*}
\centering
\caption{ASR of LiDAttack attack at different perturbation points.}
\label{2}
\begin{tabular}{c|c|c|ccc|ccc|ccc} 
\hline
\multirow{2}{*}{Dataset}   & \multirow{2}{*}{Model}       & \multirow{2}{*}{\begin{tabular}[c]{@{}c@{}}Perturbation Points\\(Perturbation Points/All Points)\end{tabular}} & \multicolumn{3}{c|}{Car} & \multicolumn{3}{c|}{Pedestrian} & \multicolumn{3}{c}{Cyclist}  \\
                           &                              &                                                                                                                & Easy & Moderate & Hard   & Easy & Moderate & Hard          & Easy & Moderate & Hard       \\ 
\hline\hline
\multirow{15}{*}{KITTI}    & \multirow{5}{*}{PointRCNN}   & 5~ (0.031\%)                                                                                                   & 0.70 & 0.83     & 0.89   & 0.73 & 0.84     & 0.90           & 0.78 & 0.86     & 0.92       \\
                           &                              & 10 (0.061\%)                                                                                                   & 0.83 & 0.91     & 0.95   & 0.85 & 0.92     & 1.00          & 0.92 & 0.94     & 1.00       \\
                           &                              & 20 (0.122\%)                                                                                                   & 0.90 & 0.95     & 0.98   & 0.93 & 0.95     & 1.00          & 0.96 & 0.96     & 1.00       \\
                           &                              & 40 (0.242\%)                                                                                                   & 0.98 & 0.98     & 1.00   & 0.98 & 1.00     & 1.00          & 0.99 & 1.00     & 1.00       \\
                           &                              & Average                                                                                                        & 0.85 & 0.92     & 0.96   & 0.87 & 0.93     & 0.98          & 0.91 & 0.94     & 0.98       \\ 
\cline{2-12}
                           & \multirow{5}{*}{PointPillar} & 5~ (0.008\%)                                                                                                   & 0.44 & 0.71     & 0.78   & 0.48 & 0.72     & 0.85          & 0.50 & 0.82     & 0.95       \\
                           &                              & 10 (0.016\%)                                                                                                   & 0.82 & 0.87     & 0.92   & 0.85 & 0.86     & 0.90          & 0.88 & 0.98     & 0.98       \\
                           &                              & 20 (0.032\%)                                                                                                   & 0.93 & 1.00     & 1.00   & 0.9  & 0.94     & 0.97          & 0.95 & 0.99     & 1.00       \\
                           &                              & 40 (0.064\%)                                                                                                   & 0.99 & 1.00     & 1.00   & 0.98 & 1.00     & 1.00          & 1.00 & 1.00     & 1.00       \\
                           &                              & Average                                                                                                        & 0.80 & 0.90     & 0.93   & 0.80 & 0.88     & 0.93          & 0.83 & 0.95     & 0.98       \\ 
\cline{2-12}
                           & \multirow{5}{*}{PV-RCNN++}   & 5~ (0.008\%)                                                                                                   & 0.23 & 0.32     & 0.58   & 0.25 & 0.44     & 0.61          & 0.43 & 0.66     & 0.87       \\
                           &                              & 10 (0.016\%)                                                                                                   & 0.55 & 0.73     & 0.89   & 0.57 & 0.64     & 0.77          & 0.75 & 0.86     & 0.89       \\
                           &                              & 20 (0.032\%)                                                                                                   & 0.64 & 0.87     & 0.93   & 0.61 & 0.73     & 0.88          & 0.85 & 0.93     & 0.95       \\
                           &                              & 40 (0.064\%)                                                                                                   & 0.80 & 0.91     & 0.96   & 0.82 & 0.93     & 0.95          & 0.90 & 0.97     & 0.99       \\
                           &                              & Average                                                                                                        & 0.56 & 0.71     & 0.84   & 0.56 & 0.69     & 0.80          & 0.73 & 0.86     & 0.93       \\ 
\hline
\multirow{15}{*}{nuScenes} & \multirow{5}{*}{PointRCNN}   & 5~(0.031\%)                                                                                                    & 0.78 & 0.85     & 0.90   & 0.80 & 0.87     & 0.95          & 0.83 & 0.87     & 0.94       \\
                           &                              & 10(0.061\%)                                                                                                    & 0.85 & 0.93     & 0.96   & 0.87 & 0.90     & 1.00          & 0.95 & 0.96     & 0.98       \\
                           &                              & 20 (0.122\%)                                                                                                   & 0.92 & 0.96     & 0.99   & 0.93 & 1.00     & 1.00          & 0.96 & 0.98     & 0.98       \\
                           &                              & 40 (0.242\%)                                                                                                   & 0.98 & 1.00     & 1.00   & 1.00 & 1.00     & 1.00          & 1.00 & 1.00     & 1.00       \\
                           &                              & Average                                                                                                        & 0.88 & 0.94     & 0.96   & 0.90 & 0.94     & 0.99          & 0.94 & 0.95     & 0.98       \\ 
\cline{2-12}
                           & \multirow{5}{*}{PointPillar} & 5(0.014\%)                                                                                                     & 0.74 & 0.90     & 0.95   & 0.75 & 0.88     & 0.94          & 0.80 & 0.93     & 0.97       \\
                           &                              & 10(0.028\%)                                                                                                    & 0.86 & 0.93     & 0.99   & 0.90 & 0.95     & 1.00          & 0.91 & 0.98     & 1.00       \\
                           &                              & 20(0.057\%)                                                                                                    & 0.95 & 0.97     & 1.00   & 0.97 & 1.00     & 1.00          & 0.97 & 1.00     & 1.00       \\
                           &                              & 40(0.115\%)                                                                                                    & 1.00 & 1.00     & 1.00   & 1.00 & 1.00     & 1.00          & 1.00 & 1.00     & 1.00       \\
                           &                              & Average                                                                                                        & 0.89 & 0.95     & 0.99   & 0.91 & 0.96     & 0.99          & 0.92 & 0.98     & 0.99       \\ 
\cline{2-12}
                           & \multirow{5}{*}{PV-RCNN++}   & 5(0.014\%)                                                                                                     & 0.48 & 0.68     & 0.85   & 0.62 & 0.70     & 0.73          & 0.69 & 0.84     & 0.88       \\
                           &                              & 10(0.028\%)                                                                                                    & 0.60 & 0.85     & 0.90   & 0.66 & 0.74     & 0.87          & 0.80 & 0.91     & 0.92       \\
                           &                              & 20(0.057\%)                                                                                                    & 0.80 & 0.90     & 0.93   & 0.80 & 0.92     & 0.95          & 0.83 & 0.96     & 0.98       \\
                           &                              & 40(0.115\%)                                                                                                    & 0.85 & 0.95     & 1.00   & 0.96 & 0.98     & 1.00          & 0.98 & 1.00     & 1.00       \\
                           &                              & Average                                                                                                        & 0.68 & 0.85     & 0.92   & 0.76 & 0.84     & 0.89          & 0.83 & 0.93     & 0.95       \\
\hline
\end{tabular}
\end{table*}

\begin{table}
\centering
\caption{ASR of LiDAttack on self-constructed dataset on different object detection models.}
\label{selft}
\begin{tabular}{c|c|c|c} 
\hline
Model                        & Perturbation Points & Pedestrian & Cyclist  \\ 
\hline\hline
\multirow{5}{*}{PointRCNN}   & 5(0.031\%)          & 0.51       & 0.61     \\
                             & 10(0.061\%)         & 0.64       & 0.67     \\
                             & 20(0.122\%)         & 0.82       & 0.87     \\
                             & 40(0.242\%)         & 0.90       & 0.93     \\
                             & Average             & 0.72       & 0.77     \\ 
\hline
\multirow{5}{*}{PointPillar} & 5(0.002\%)          & 0.32       & 0.30~    \\
                             & 10(0.003\%)         & 0.47       & 0.52     \\
                             & 20(0.007\%)         & 0.68       & 0.75     \\
                             & 40(0.013\%)         & 0.90       & 0.94     \\
                             & Average             & 0.59       & 0.63     \\ 
\hline
\multirow{5}{*}{PV-RCNN++}   & 5(0.002\%)          & 0.12       & ~0.20~   \\
                             & 10(0.003\%)         & 0.32       & ~0.40~   \\
                             & 20(0.007\%)         & 0.56       & 0.61     \\
                             & 40(0.013\%)         & 0.75       & 0.78     \\
                             & Average             & 0.44       & 0.50     \\
\hline
\end{tabular}
\end{table}

\textbf{Different object.} LiDAttack is effective against different objects. Table \ref{2} shows the ASR of LiDAttack's attack on different object detection models, different number of perturbation points, different types of objects and different recognition difficulty. Fig. \ref{different object and points}(a) shows the ASR of LiDAttack attack for different object detection models and different objects when the number of perturbation points is 10. It can be observed that for ``Car", ``Pedestrian" and ``Cyclist" three types of target objects, LiDAttack can effectively attack them all, For PointRCNN and PointPillar, the ASR of LiDAttack on different objects is more than 80\%, and the ASR of PV-RCNN++ is slightly lower, which proves the PV-RCNN++ effectiveness and stability from the side. Moreover, the LiDAttack attack is effective for ``Cyclist'' type of attack always has the greatest ASR, which may be due to the fact that this type of target object has a smaller number of point clouds.

\textbf{Number of perturbation points.} LiDAttack requires only a small number of perturbation points for a successful attack on an object detection model. Fig. \ref{different object and points}(b) shows the ASR of LiDAttack on PointRCNN for different perturbation points for objects with different recognition difficulty. As the number of perturbation points increases, the ASR also increases. When the number of perturbation points reaches 40, the ASR of LiDAttack on objects with different recognition difficulty is close to 100\%. This indicates that only a small number of perturbation points need to be added to the target object to realize a successful attack.

\textbf{Different recognition difficulty.} LiDAttack's ASR for ``Easy" recognition difficulty are lower than those for ``Moderate" and ``Hard".
Fig. \ref{different object and points}(b) shows the ASRs on the target object ``Car" at different perturbation points with different recognition difficulty. From Fig. \ref{different object and points}(b), it can be found that the ASRs of LiDAttack on the ``Easy" class is lower compared to ``Moderate" and ``Hard" is lower due to the fact that the class ``Easy"  occlusion level is fully visible, and the maximum truncation is 15\%, so the ``Easy" has a higher number of point clouds, and when a small number of perturbed points are added the harder it is to attack.

\begin{figure}
\centering
\includegraphics[width=0.48\textwidth]{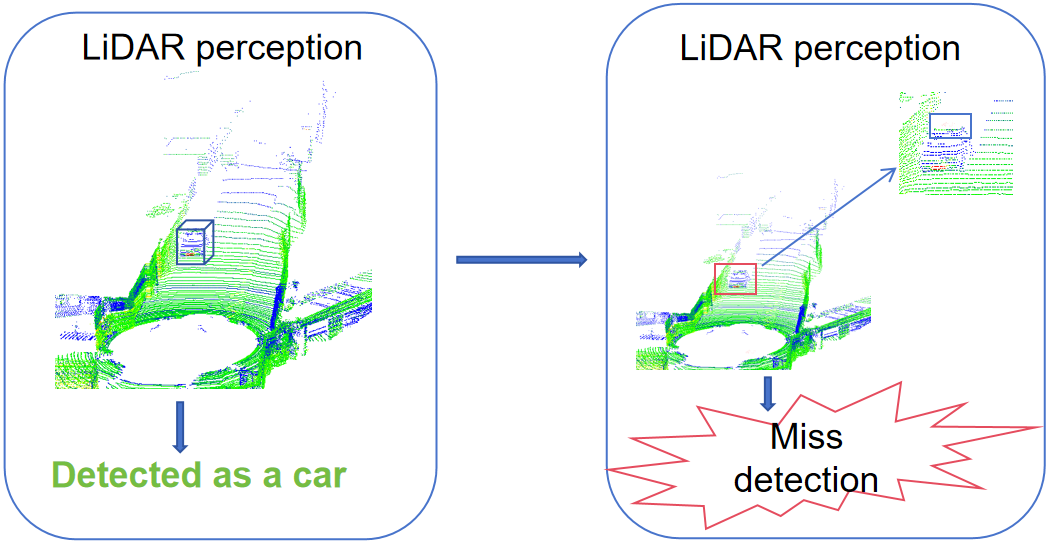}
\caption{The comparison between the unattacked point cloud and the point cloud after the attack on the digital domain.}
\label{attackbeforeandafter}
\end{figure}

The comparison between the unattacked point cloud and the point cloud after the attack on the digital domain is given in Fig. \ref{attackbeforeandafter}, from which it can be seen that the perturbed points are hardly visible after adding the perturbation.

\begin{framed}
\textbf{Answer to RQ1:} LiDAttack has a satisfying performance in terms of attack effectiveness. The ASR increases significantly with the increase of the number of perturbation points, and a high ASR can be achieved with only 10 perturbation points, especially when the number of perturbation points is close to 40, the ASR achieved by LiDAttack reaches almost 100\% for three different object detection models.

\end{framed}

\begin{figure}
\centering
\includegraphics[width=0.48\textwidth]{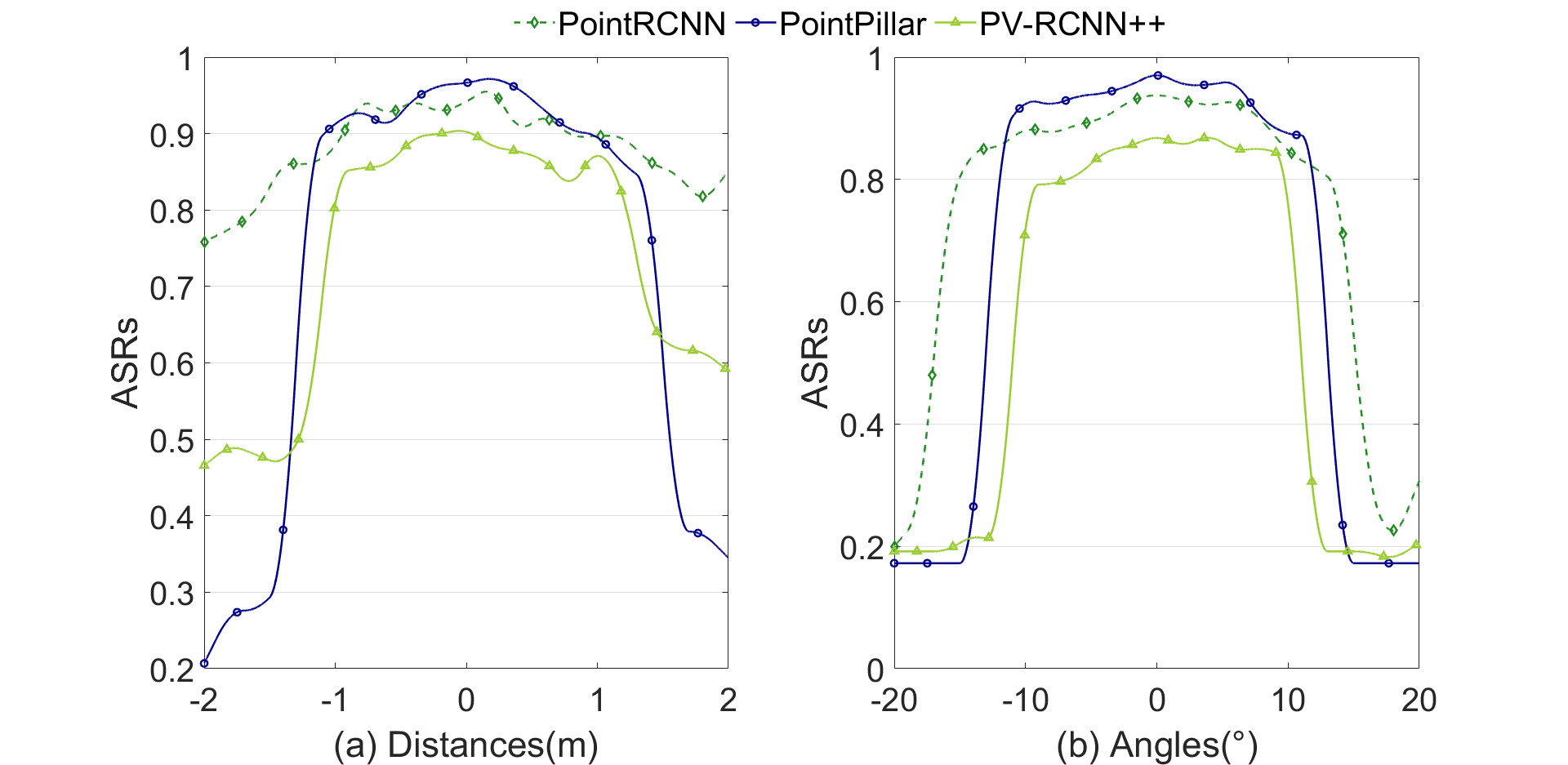}
\caption{ASR of LiDAttack attack under different distances and angles. (a) ASR of LiDAttack attack under different distances. (b) ASR of LiDAttack attack under different angles.}
\label{different distance and angle}
\end{figure}

\subsection{\textbf{RQ2.} How about LiDAttack's robustness?}

\begin{figure}
\centering
\includegraphics[width=0.48\textwidth]{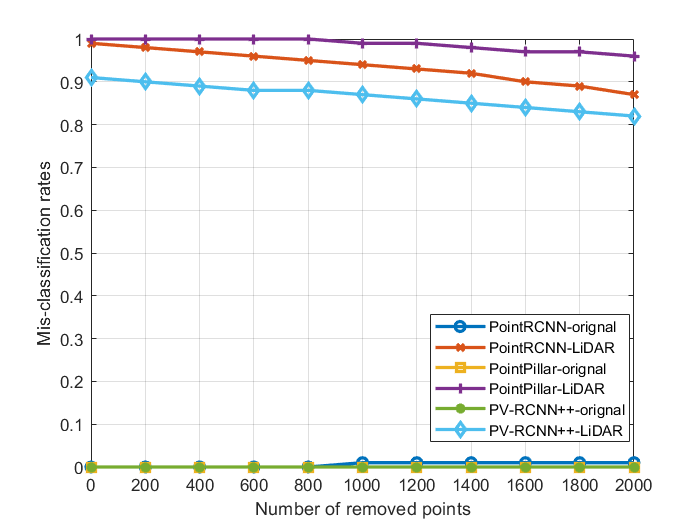}
\caption{The misclassification rate of the original model for target objects and the misclassification rate in the presence of the LiDAttack attack.}
\label{SRS}
\end{figure}

When reporting the results, we focus on the following aspects, i.e., distance and angles. The evaluation results are shown in Fig. \ref{different distance and angle} and Fig. \ref{SRS}. 

In real world scenarios, it is not possible to ensure that an object can be placed back to its original position accurately, so it is necessary to explore the potential impact of object placement bias on ASR. Specifically, we consider two main factors, i.e., the distance between the object and the LiDAR, and the angle at which the object is placed relative to the LiDAR.

\textbf{Robustness in distance change.} LiDAttack attack is robust to different object detection models over a range of distances. The distance between the object and the LiDAR can be obtained by translating the coordinates of the object. From Fig. \ref{different distance and angle}(a), LiDAttack shows strong robustness in the perturbation range of -1m to 1m. This shows that our method has better robustness compared to the adversarial objects generated in paper \cite{zhang2022towards}. However, the ASR decreases as the perturbation distance increases or decreases. In particular, the ASR against PointPillar and PV-RCNN++ declines faster than the ASR against PointRCNN. This difference may be due to the fact that PointPillar and PV-RCNN++ employ different strategies to process point cloud data.PointPillar does this by converting the point cloud into a sparse 3D voxel representation and utilizing a convolutional neural network for feature extraction and prediction, while PV-RCNN++ divides the point cloud into multiple local regions for feature extraction and prediction separately. These approaches may be more effective in capturing the local structure and contextual information of the point cloud data, which enhances the defense of the object detection model against LiDAttack.

\textbf{Robustness in angle change.} LiDAttack attack is robust to different object detection models over a range of angles. To get the point cloud at different angles, the object is rotated and scanned. The rotation matrix around the Z-axis is as follows.
\begin{equation}
R_z=\left[ \begin{matrix}
	\cos \psi&		-\sin \psi&		0\\
	\sin \psi&		\cos \psi&		0\\
	0&		0&		1\\
\end{matrix} \right]  
\label{Angles}
\end{equation}

From the Fig. \ref{different distance and angle}(b), it can be observed that LiDAttack is robust when the angle is between -10° and 10°, and the ASR then decreases sharply with the increase of the angle change, which indicates that LiDAttack is sensitive to the angle change. This sensitivity mainly stems from the fact that rotating the adversarial point cloud changes the local features, and LiDAttack needs to re-search for suitable adversarial samples at the new angle.

\textbf{Robustness in face of defense.}  Simple random sampling (SRS) destroys the structure of the point cloud by randomly deleting a portion of the points in the point cloud, thus achieving the effect of reducing the attack of the adversarial examples on the model. To demonstrate that the LiDAttack attack is also robust to defense, a random selection of (0, 2000) points is made from the point cloud, followed by pre-processing the input point cloud. The order of the input points is irrelevant and does not impact the classification performance. Randomly deleting certain points will have a certain chance of taking away the perturbation points, thus reducing the ASR. As shown in Fig. \ref{SRS}, even randomly removing up to 1.7\% (nearly 2000 points) of the points from the original point cloud does not significantly reduce the misclassification of the target detection model due to the inclusion of perturbed points generated by LiDAttack.

\begin{figure}
\centering
\includegraphics[width=0.48\textwidth]{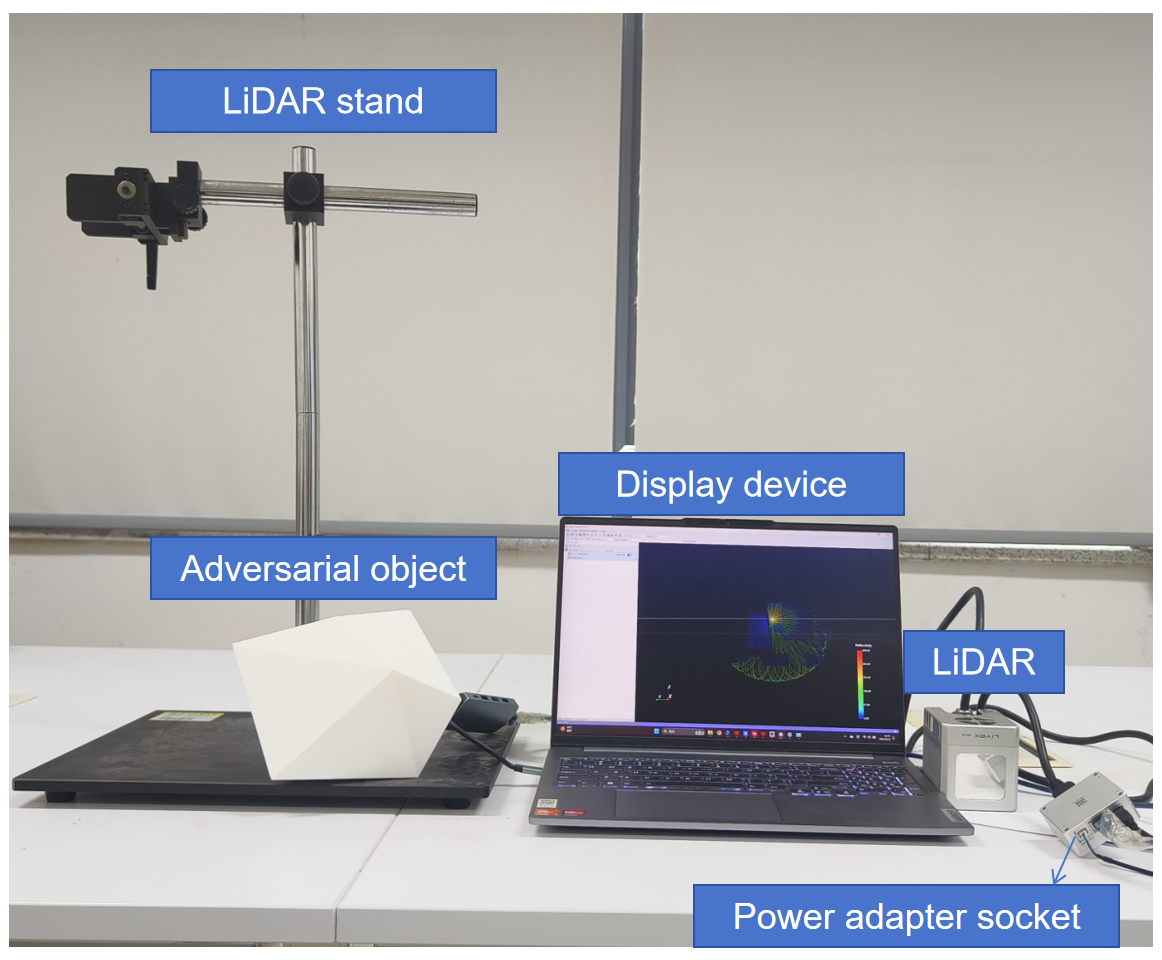}
\caption{Collection device configuration.}
\label{device}
\end{figure}

\begin{framed}
\textbf{Answer to RQ2: }LiDAttack shows high ASR and stability at distances of ±1 meters and angles of -10° to 10°, maintaining performance even when scanning at different angles and distances. When using SRS to defend against LiDAttack attacks, removing 1.7\% of the point cloud still has a small impact on the effectiveness of the attack.
\end{framed}

\subsection{\textbf{RQ3.} How does the physical Practicality of LiDAttack attack?}

\begin{figure}
\centering
\includegraphics[width=0.48\textwidth]{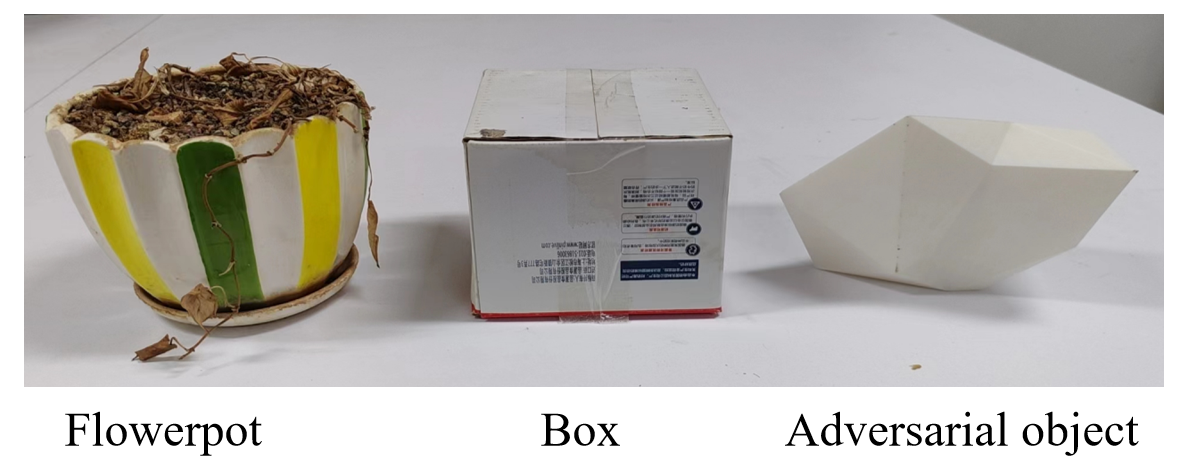}
\caption{Visualization of the flowerpot, box, perturbed object with size limited to $0.2m\times 0.2m\times 0.2m$ used in indoor experiments.}
\label{three_object}
\end{figure}

\begin{figure}
\centering
\includegraphics[width=0.48\textwidth]{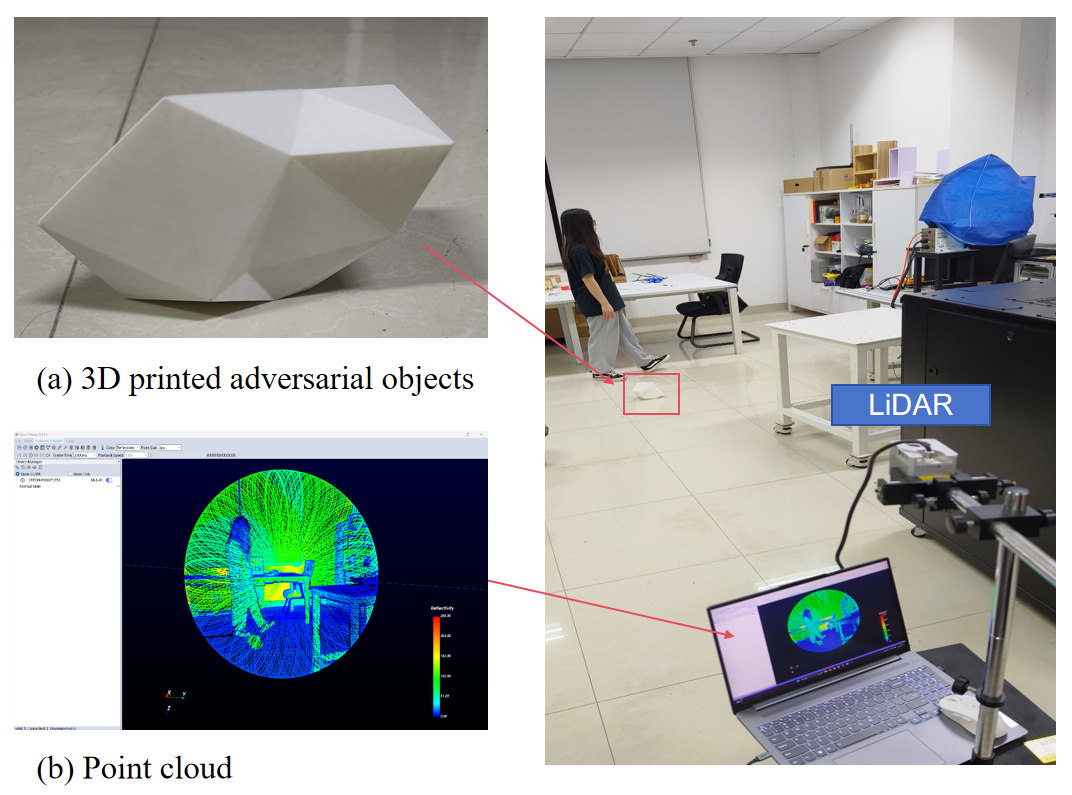}
\caption{Visualization of the flowerpot, box, perturbed object with size limited to $0.2m\times 0.2m\times 0.2m$ used in indoor experiments.}
\label{indoor}
\end{figure}

\begin{figure}
\centering
\includegraphics[width=0.48\textwidth]{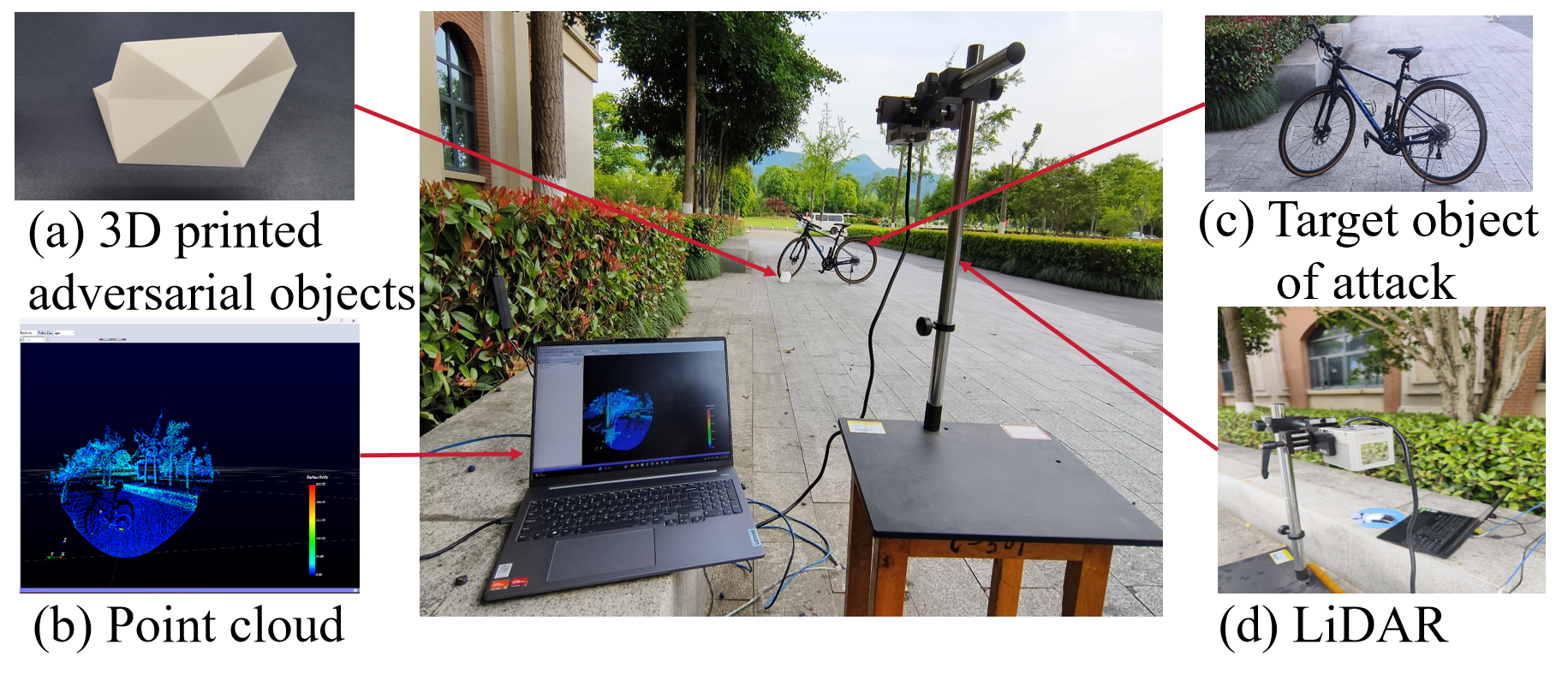}
\caption{Outdoor experiment, perturbed object
with size limited to $0.2m\times 0.2m\times 0.2m$ used in  experiments.}
\label{pyscial_attack}
\end{figure}

\begin{figure}
\centering
\includegraphics[width=0.48\textwidth]{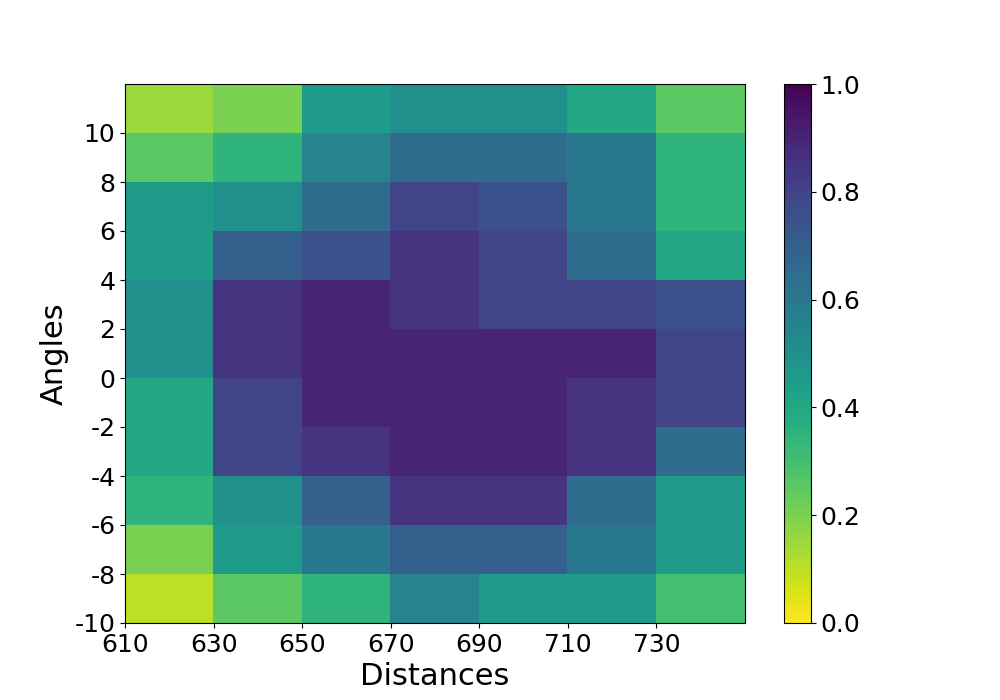}
\caption{Misidentification rate in the physical world.}
\label{CM}
\end{figure}

When reporting the results, we focus on the following aspects, i.e., indoor and outdoor. We did experiments in the physical world and the results are displayed in  Fig. \ref{device} to Fig. \ref{CM}. In order to show our experimental process more intuitively, we have placed a demonstration video of our experimental process in \url{https://github.com/Cinderyl/self-constructed-data}. 


Fig. \ref{device} illustrates the LiDAR configuration used for point cloud data acquisition. In this experiment, the attacker achieved a spoofing attack on the LiDAR by placing an adversarial scrambling object next to the target object. These interfaces include a dedicated laser detection rangefinder connector interface, a synchronisation signal interface, a power supply interface and an Ethernet interface for data communication. The LiDAR unit communicates with external devices over Ethernet using the UDP protocol. We add a negative sign to indicate when the object is closer than the original distance or when the object is rotated clockwise relative to the original position.

\textbf{Indoor experiment.} LiDAttack attack remains effective indoors. In the indoor environment, the size of the adversarial objects was limited to $0.2m\times 0.2m\times 0.2m$. The flowerpot and box were used as controls. The printed adversarial objects and normal objects are shown in Fig. \ref{three_object}. The device is built as shown in Fig. \ref{indoor}, Table \ref{three_digit} shows the misclassification rate of the object detection model, the misclassification rate shows that normal objects can not make the object detection model to identify incorrectly, but our adversarial objects can make the object detection model to identify incorrectly.

\textbf{Outdoor experiment.} LiDAttack attack remains effective outdoors. Various physical constraints were taken into account and the meshes of adversarial objects were modified to enable them to be 3D printed in the real world. We limit the size of printed objects to $0.2m\times 0.2m\times 0.2m$, Fig. \ref{pyscial_attack} shows the environment of our experiment. The printed object was placed next to the attacked bicycle, the scanned point cloud is displayed in the computer in real time, and we re-input the point cloud after adding the perturbed objects into the target detection model, and the detection results are shown in Fig. \ref{CM}. The distance between the original LiDAR and the cyclist is 670cm, and the acquisition angle is 0°. The average results of the frames collected at different ranges are reported. As can be found in Fig. \ref{CM}, ASR decreases as the distance to the object gets closer. This is due to the fact that the target vehicle generates more LiDAR points when it is closer to the LiDAR, which facilitates feature learning for the detection model. It can also be noticed from the table that the change of angle tends to decrease the ASR. More than 85\% ASR is maintained by LiDAttack in the distance range of (630, 730)cm at the angle of 0°, demonstrating its robustness over distance. When the distance is fixed at (670, 690)cm, LiDAttack can maintain the ASR more than 80\% over an angle range of (-6°, 6°), showing its robustness over angles as well. However, this aggressiveness is only effective over a small angular range, as changes in angle result in large differences in the point cloud of the adversarial object scanned by the LiDAttack. In contrast, changes in distance have less of an effect on the point cloud.

\begin{table}
\centering
\caption{Misclassification rate of flowerpot, box and adversarial object in indoor scenario.}
\label{three_digit}
\begin{tblr}{
  cells = {c},
  vline{2} = {-}{},
  hline{1-2,5} = {-}{},
}
\diagbox{Object}{Model} & PointRCNN & PointPillar & PV-RCNN++ \\
{Flowerpot}       & 0.15      & 0.10        & 0.00      \\
{Box}             & 0.05      & 0.15        & 0.05      \\
{LiDAttack}       & 0.90      & 0.90        & 0.75      
\end{tblr}
\end{table}

\begin{framed}
\textbf{Answer to RQ3:} In indoor experiments, LiDAttack can effectively mislead the object detection model, while normal objects cannot. In the outdoor experiments, LiDAttack can maintain more than 85\% ASR at a distance range of (630, 730) cm, while it can maintain more than 80\% ASR at a distance of (670, 690) cm and an angle range of (-6°, 6°). This shows that LiDAttack has strong effectiveness and robustness under different experimental conditions.

\end{framed}

\subsection{\textbf{RQ4}. How to defend against LiDAttack and improve model robustness?}

The adversarial examples generated by LiDAttack are used to enhance the defense capability of the target detection model through adversarial training. The core idea of adversarial training is to introduce adversarial examples during the training process so that the model gradually adapts and defends against adversarial attack during the learning process. Specifically, adversarial examples are generated by LiDAttack and the proportion of adversarial samples is gradually increased in the training dataset starting from using all original samples until further increase in the proportion of adversarial samples does not significantly improve the model performance.





As shown in Table. \ref{training}, the models PointRCNN and PointPillar exhibited a notable performance degradation when subjected to the LiDAttack, with an increase in the misidentification rate of 70\% to 80\%. This suggests that the LiDAttack has a significantly adverse impact on these models, effectively impairing their detection capabilities. On the other hand, although the PV-RCNN++ was also affected by the attack, the increase in its misidentification rate was relatively smaller, at 60\% to 70\%. This indicates that PV-RCNN++ has certain structural advantages or higher robustness in its design, enabling it to better resist a certain level of attack. To counter such attack, adversarial training was employed as a defensive mechanism. After adversarial training, when the models were attacked again with LiDAttack, PointRCNN's misidentification rate significantly decreased by 60\% to 70\%. This indicates that adversarial training effectively enhanced PointRCNN's resistance to the LiDAttack. Similarly, the misidentification rates of PointPillar and PV-RCNN++ also declined after adversarial training when faced with LiDAttack, decreasing by 50\% to 60\% respectively. These results confirm the effectiveness of adversarial training as a defensive measure.

\begin{framed}
\textbf{Answer to RQ4:} Adversarial training greatly improves the model's ability to defend against LiDAttack attack, with ASR dropping by about 60\%.
\end{framed}

\begin{table}
\setlength{\tabcolsep}{3pt}
\centering
\caption{misidentification rate of LiDAttack after defense using adversarial training.}
\label{training}
\begin{tabular}{c|ccc|ccc|ccc} 
\hline
\multirow{2}{*}{}                                              & \multicolumn{3}{c|}{PointRCNN} & \multicolumn{3}{c|}{PointPillar} & \multicolumn{3}{c}{PV-RCNN++}  \\
                                                               & Easy & Mod. & Hard             & Easy & Mod. & Hard               & Easy & Mod. & Hard             \\ 
\hline\hline
Orignal                                                        & 0.04 & 0.08 & 0.13             & 0.05 & 0.08 & 0.12               & 0.01 & 0.04 & 0.07             \\
LiDAttack                                                      & 0.87 & 0.91 & 0.98             & 0.85 & 0.90 & 0.93               & 0.62 & 0.74 & 0.85             \\
\begin{tabular}[c]{@{}c@{}}Adv\_train\end{tabular} & 0.15 & 0.24 & 0.31             & 0.22 & 0.38 & 0.37               & 0.08 & 0.17 & 0.21             \\
\hline
\end{tabular}
\end{table}
\section{Conclusion}
A novel black-box attack  LiDAttack is proposed, which aims to attack the target detection model in a LiDAR-based sensing system. The method uses a genetic simulated annealing algorithm to optimise the position of the perturbation point to achieve stealth and robustness while ensuring the effectiveness of the attack. Experimental results show that LiDAttack exhibits excellent performance against mainstream target detection models such as PointRCNN, PointPillar, and PV-RCNN++ on KITTI, nuScenes, and self-constructed data datasets, and effectively improves the robustness of the models through adversarial training.

However, LiDAttack has some limitations. Firstly, LiDAttack's ASR decreases significantly when the angle changes, which limits its application in complex scenarios. Secondly, the adversarial training can only partially improve the robustness of the model, and there still exists the possibility of successful attack, which requires further research on more comprehensive defense methods.

In order to further improve the attack effect and application scope of LiDAttack, future research will be conducted in the following aspects, i.e.,  1) exploring more effective local search strategies, e.g., using methods such as reinforcement learning, to improve ASR; 2) researching more comprehensive adversarial training, e.g., using meta-learning or data augmentation, to improve the robustness of the model; 3) applying LiDAttack to other scenarios, such as drones, robots, etc., to explore its application potential in more fields.

\bibliographystyle{IEEEtran}

\begin{IEEEbiography}[{\includegraphics[width=1.0in,height=1.25in,clip,keepaspectratio]{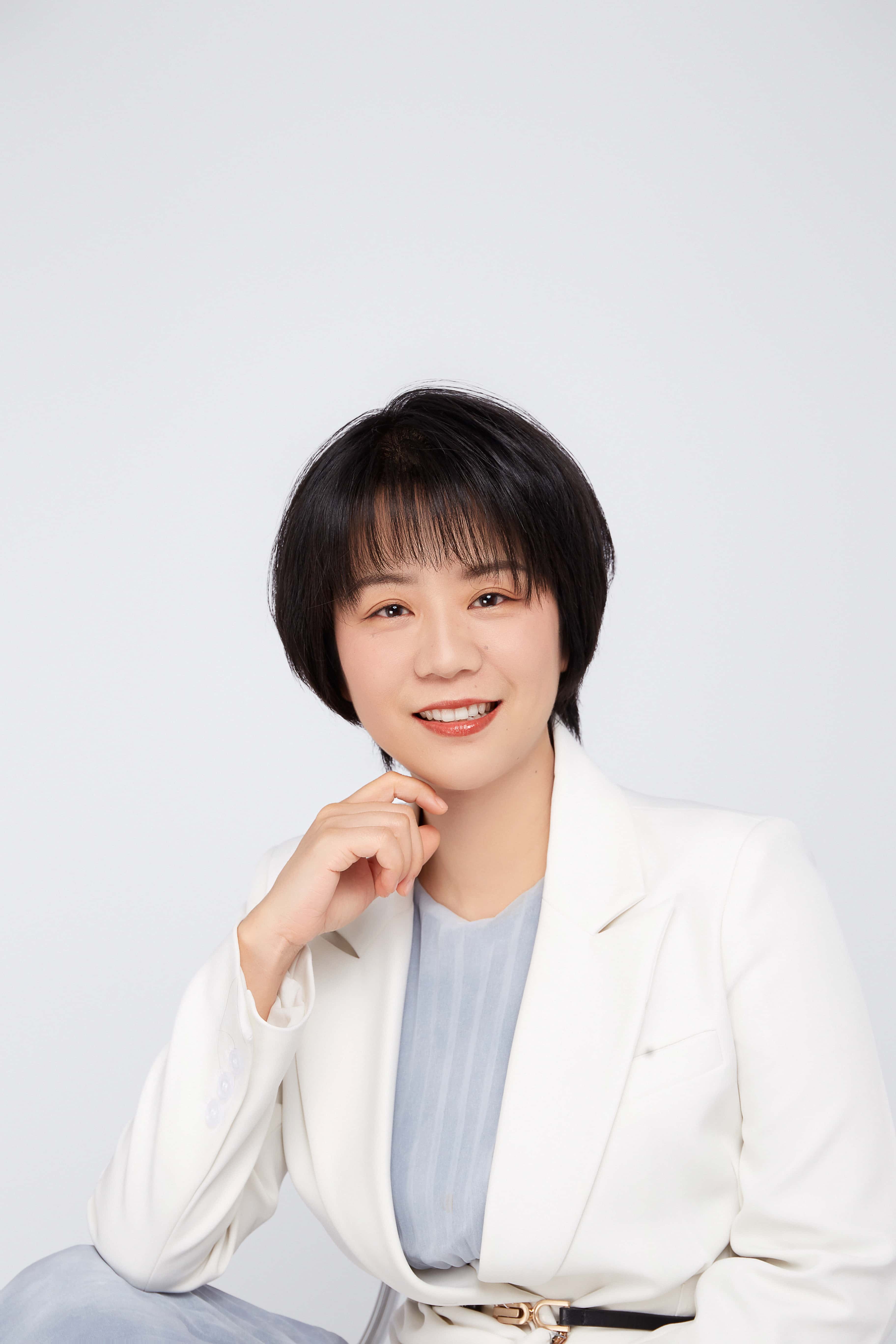}}]{Jinyin Chen}
		received BS and PhD degrees from
	Zhejiang University of Technology, Hangzhou,
	China, in 2004 and 2009, respectively.
	
	She studied
	evolutionary computing in Ashikaga Institute
	of Technology, Japan in 2005 and 2006.
	She is currently an Associate Professor with the
	Zhejiang University of Technology, Hangzhou,
	China. Her research interests include artificial
	intelligence security, graph data mining and evolutionary
	computing.
	\end{IEEEbiography}
 \begin{IEEEbiography}[{\includegraphics[width=1.0in,height=1.25in,clip,keepaspectratio]{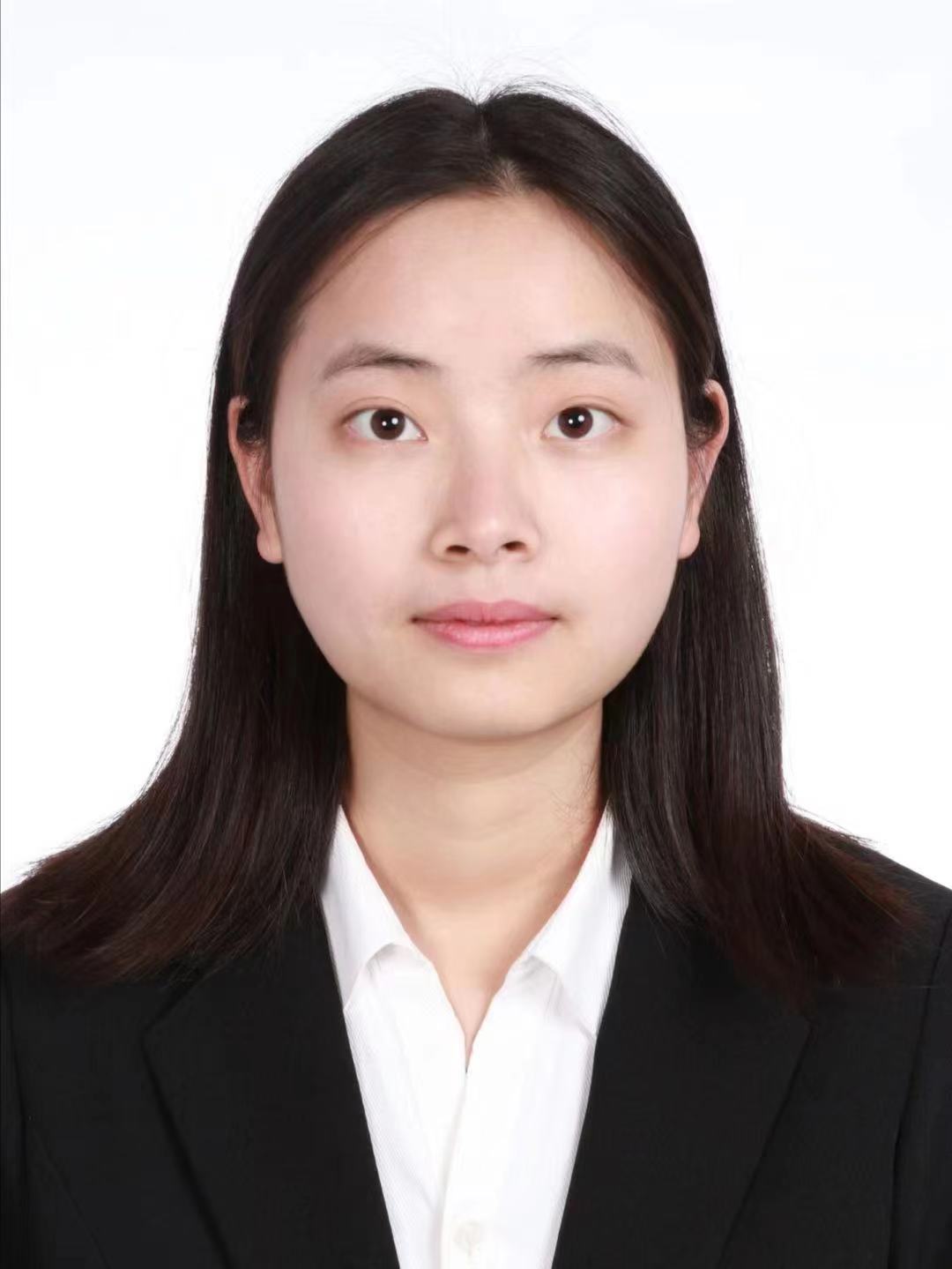}}]{Danxin Liao}
		is currently pursuing the master degree with the college of Information Engineering, Zhejiang University of Technology.

       Her research interests include deep learning and artificial intelligence security.

	\end{IEEEbiography}
  
  \begin{IEEEbiography}[{\includegraphics[width=1.0in,height=1.25in,clip,keepaspectratio]{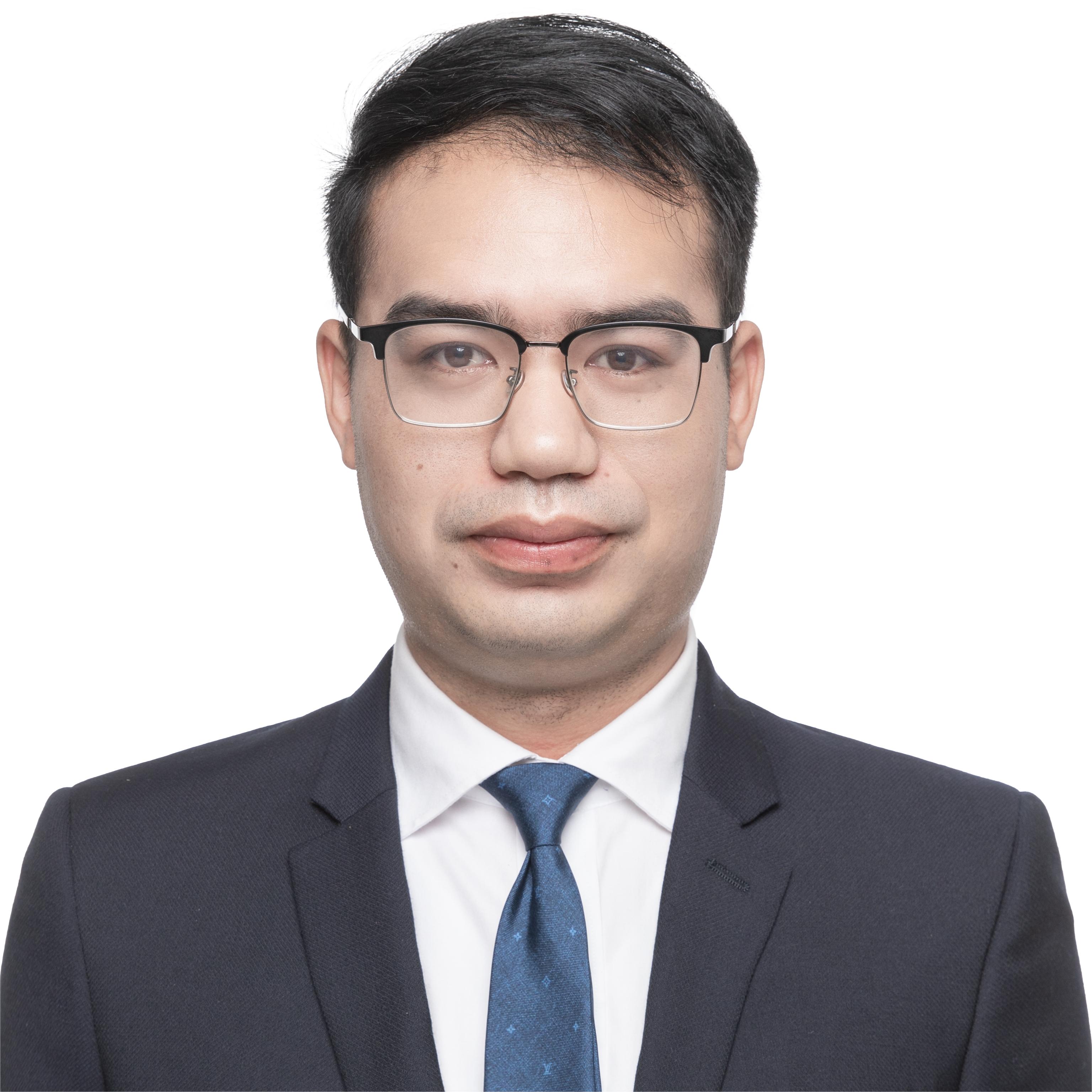}
 }]{Sheng Xiang}
		received M.S. and Ph.D. degrees from Grauate School of Information Science and Technology from Hokkaido University, Hokkaido, Japan, in 2018 and 2021, respectively. He is currently a university lecturer at the Information Science and Technology College, Zhejiang University of Technology. 

        His research interests include deep learning and machine vision.

	\end{IEEEbiography}
	\begin{IEEEbiography}[{\includegraphics[width=1in,height=1.25in,clip,keepaspectratio]{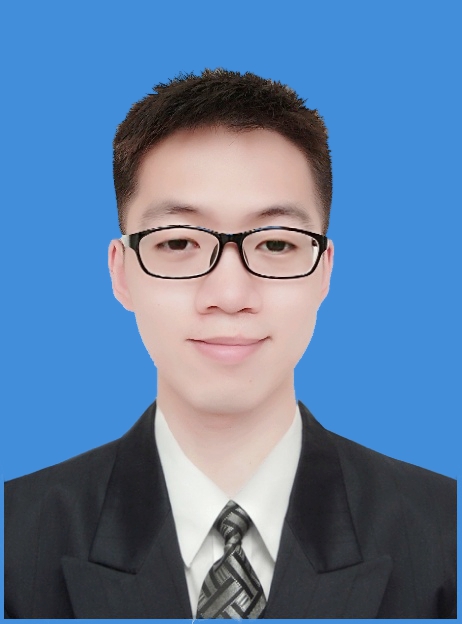}}]{Haibin Zheng}
		received B.S. and Ph.D. degrees from Zhejiang University of Technology, Hangzhou, China, in 2017 and 2022, respectively. He is currently a university lecturer at the Institute of Cyberspace Security, Zhejiang University of Technology. 
		
		His research interests include deep learning and artificial intelligence security.
	\end{IEEEbiography}


\end{CJK}
\end{document}